\documentclass[10pt,a4paper]{article}
\usepackage{fix-cm}
\usepackage{fontspec}
\usepackage[english]{babel}
\newfontfamily\magictitlefont{texgyreheros-regular.otf}[
    BoldFont=texgyreheros-bold.otf,
    ItalicFont=texgyreheros-italic.otf,
    BoldItalicFont=texgyreheros-bolditalic.otf, Ligatures=TeX]
\usepackage[a4paper,margin=2.45cm]{geometry}
\usepackage{microtype}
\usepackage{setspace}
\usepackage{graphicx}
\usepackage[normalem]{ulem}
\usepackage{float}
\graphicspath{{figures/}}
\usepackage{subcaption}
\usepackage{booktabs}
\usepackage{array}
\usepackage{multirow}
\usepackage{tabularx}
\usepackage{makecell}
\usepackage{amsmath,amssymb,bm,mathtools}
\usepackage{siunitx}

\usepackage{xcolor}
\usepackage[most]{tcolorbox}
\definecolor{MagicBlue}{HTML}{2D68D8}
\definecolor{MagicAbstractBg}{HTML}{F2FAFE}
\definecolor{MagicAbstractBorder}{HTML}{C3E2F2}
\definecolor{DraftBlue}{HTML}{3568A8}
\definecolor{RuleGray}{HTML}{D7DCE3}
\usepackage{enumitem}
\setlist{nosep,leftmargin=1.6em}
\usepackage[numbers,sort&compress]{natbib}
\usepackage{ragged2e}
\usepackage{xurl}
\usepackage[titles]{tocloft}
\cftsetindents{section}{0em}{1em}
\cftsetindents{subsection}{1em}{1.7em}
\cftsetindents{subsubsection}{2.7em}{2.4em}

\usepackage[colorlinks=true,linkcolor=MagicBlue,citecolor=MagicBlue,urlcolor=MagicBlue]{hyperref}
\usepackage[nameinlink,noabbrev]{cleveref}

\newcommand{\tablefont}{\fontsize{8.5}{11}\selectfont}
\renewcommand{\arraystretch}{1.12}
\AtBeginDocument{%
    \setlength{\abovedisplayskip}{6pt plus 2pt minus 1pt}%
    \setlength{\belowdisplayskip}{6pt plus 2pt minus 1pt}%
    \setlength{\abovedisplayshortskip}{4pt plus 1pt minus 1pt}%
    \setlength{\belowdisplayshortskip}{4pt plus 1pt minus 1pt}%
}

\usepackage{titlesec}
\titleformat{\section}{\normalfont\color{black}\fontsize{12}{14.4}\selectfont\bfseries}{\thesection}{0.75em}{}
\titleformat{\subsection}{\normalfont\fontsize{11}{13.2}\selectfont\bfseries}{\thesubsection}{0.75em}{}
\titleformat{\subsubsection}{\normalfont\fontsize{10}{12}\selectfont\bfseries}{\thesubsubsection}{0.75em}{}
\titleformat{\paragraph}[runin]{\normalfont\bfseries}{}{0pt}{}
\titlespacing*{\section}{0pt}{18pt plus 2pt minus 2pt}{8pt}
\titlespacing*{\subsection}{0pt}{14pt plus 2pt minus 1pt}{6pt}
\titlespacing*{\subsubsection}{0pt}{10pt plus 1pt minus 1pt}{4pt}
\titlespacing*{\paragraph}{0pt}{4pt}{0.5em}
\newtcolorbox{magicabstractbox}{
    enhanced,colback=MagicAbstractBg,colframe=MagicAbstractBorder,
    boxrule=0.45pt,arc=4mm,
    left=15pt,right=15pt,top=12pt,bottom=13pt,
    before skip=48pt,after skip=0pt
}
\renewenvironment{abstract}{%
    \begin{magicabstractbox}%
    \rmfamily\fontsize{10}{12}\selectfont\setstretch{1.1}%
    \setlength{\parskip}{0.18em}%
    {\centering\rmfamily\bfseries\fontsize{12}{14.4}\selectfont\color{black} Abstract\par}\vspace{6pt}\noindent\ignorespaces
}{\end{magicabstractbox}}

\crefname{section}{Section}{Sections}
\crefname{subsection}{Section}{Sections}
\crefname{figure}{Figure}{Figures}
\crefname{table}{Table}{Tables}
\crefname{equation}{Equation}{Equations}
\usepackage{tikz}
\usetikzlibrary{arrows.meta,decorations.pathreplacing}
\usepackage{needspace}
\usepackage[section]{placeins}
\usepackage{etoolbox}
\AtBeginEnvironment{thebibliography}{\RaggedRight}
\apptocmd{\bibsection}{\phantomsection\addcontentsline{toc}{section}{\refname}}{}{}
\pretocmd{\section}{\Needspace{5\baselineskip}}{}{}
\pretocmd{\subsection}{\Needspace{4\baselineskip}}{}{}
\pretocmd{\subsubsection}{\Needspace{3\baselineskip}}{}{}
\hypersetup{pdftitle={Magic-W0: A Structured World--Action Foundation Model for Physical Intelligence},pdfauthor={Magic-Lab Team, Magiclab Robotics Inc.}}

\usepackage{xspace}
\newcommand{\model}{Magic-W0\xspace}
\newcommand{\modelFullTitle}{Magic-W0: A Structured World--Action Foundation Model for Physical Intelligence}
\newcommand{\VLA}{VLA\xspace}
\newcommand{\WAM}{WAM\xspace}

\newcommand{\obs}{\mathbf{o}}
\newcommand{\lang}{\mathbf{l}}
\newcommand{\state}{\mathbf{s}}
\newcommand{\action}{\mathbf{a}}

\newcommand{\Loss}{\mathcal{L}}

\title{\modelFullTitle}
\author{Magic-Lab Team, Magiclab Robotics Inc.}
\date{September 2026}

\begin{document}
\AddToHookNext{shipout/foreground}{%
    \begin{tikzpicture}[remember picture,overlay]
        \node[anchor=north west,inner sep=0pt,xshift=2.45cm,yshift=-0.95cm]
            at (current page.north west)
            {\includegraphics[width=3.1cm]{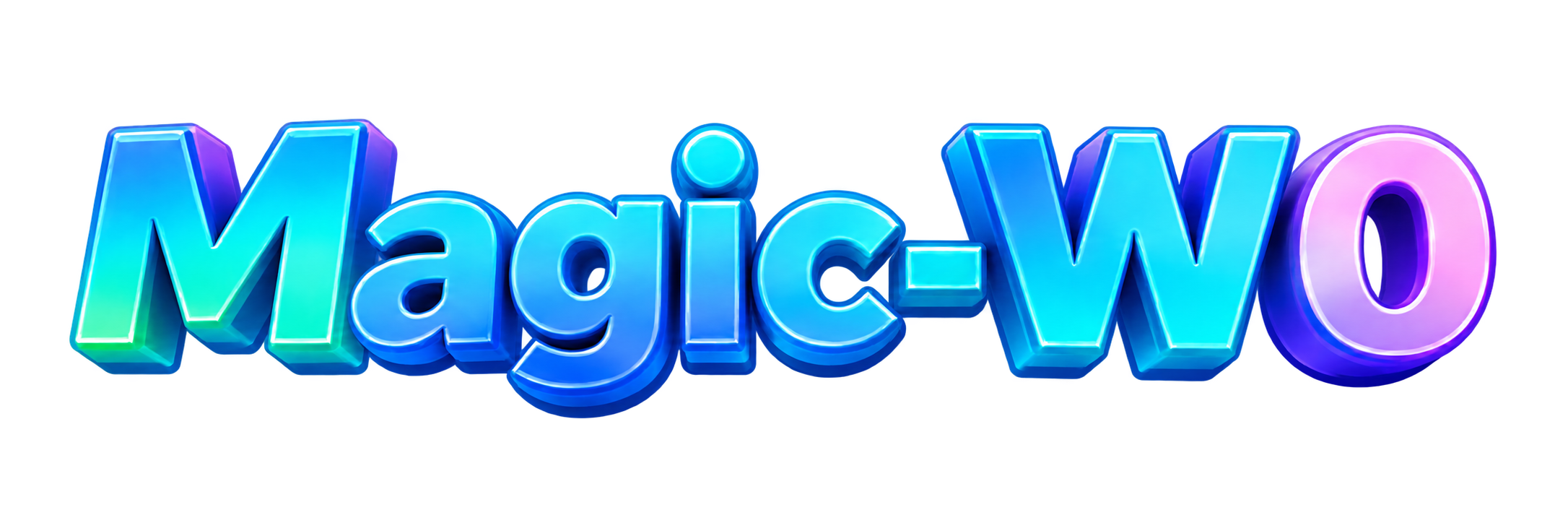}};
        \draw[RuleGray,line width=0.45pt]
            ([xshift=2.45cm,yshift=-1.79cm]current page.north west) --
            ([xshift=-2.45cm,yshift=-1.79cm]current page.north east);
    \end{tikzpicture}%
}
\vspace*{-18pt}
\begin{center}
{\setstretch{1.0}\magictitlefont\fontsize{16}{20}\selectfont\bfseries%
Magic-W0: A Structured World--Action\\
Foundation Model for Physical Intelligence\par}
\vspace{1.0em}
{\large Magic-Lab Team, Magiclab Robotics Inc.\par}
\vspace{0.55em}
{\small\color{DraftBlue} September 2026\par}
\vspace{0.2em}
\end{center}
\thispagestyle{plain}
\begin{abstract}
\noindent
World--action models (WAMs) incorporate physical world modeling into robot policies by learning action-conditioned environment dynamics. Existing WAMs primarily rely on future observation reconstruction or generic latent state prediction, but still lack structured, control-oriented representations of the physical world and a unified modeling framework tightly coupled with action generation. We propose \textbf{Magic-W0}, a world--action foundation model that jointly models structured physical state evolution and continuous action generation. To model physical state evolution explicitly, we introduce \textbf{Structured World Transition}, which organizes environment evolution during robot interaction into \textbf{Current State--Transition--Future State}. The Current State combines vision--language model (VLM) context with Current 3D Geometry. The Transition is captured by 3D Motion, which represents action-induced three-dimensional state changes. Future Semantics describes task-relevant changes in future observations, providing a semantic representation of the Future State. To unify world prediction and action generation, Magic-W0 introduces a layer-aligned world--action interaction architecture. Action hypotheses condition future world-transition prediction, realizing \textbf{action-conditioned world transition}, while predicted world representations in turn inform action generation, realizing \textbf{world-informed action generation}. This bidirectional interaction tightly couples physical world modeling with continuous action generation. Magic-W0 is pre-trained at scale on egocentric human manipulation data, Universal Manipulation Interface (UMI) data, real-robot trajectories, and simulation data. Geometry, 3D motion, and future semantic representations are learned under latent supervision from pre-trained visual models, jointly supporting structured world modeling and action generation. Inference-time interventions further reveal that structured world representations respond to changes in candidate actions, while action-related information propagates through shared 3D representations into future semantic predictions. On RoboDojo-Sim, Magic-W0 achieves an average Score of 36.75, ranking first among all compared methods. On multiple real-robot tasks, Magic-W0 achieves strong performance after fine-tuning with limited downstream data, demonstrating generalization and rapid adaptation in complex embodied tasks.
\par\smallskip
\noindent\textbf{Project page:} \url{https://embodied.magiclab.top/works/wam/magic-w0/index.html}
\par\noindent\textbf{GitHub:} \url{https://github.com/MagiclabRobotics/Magic-W0}
\end{abstract}

\clearpage
\begingroup
\setstretch{1.0}
\hypersetup{linktoc=all}
\pdfbookmark[1]{Contents}{contents}
\tableofcontents
\endgroup
\clearpage
\section{Introduction}
\label{sec:introduction}

Developing general-purpose robot policies that understand open-ended language instructions, perceive complex environments, and execute continuous control is a major goal of embodied intelligence.
Vision--language--action (VLA) models combine vision--language pre-training with robot trajectory learning.
This allows policies to use large-scale visual semantic knowledge to understand tasks and generate actions from visual observations, language instructions, and proprioceptive states.
RT-1 and RT-2 demonstrated the value of large-scale robot data and internet vision--language knowledge for task coverage and instruction generalization~\citep{brohan2022rt1,zitkovich2023rt2}.
Open X-Embodiment, Octo, and OpenVLA subsequently advanced general-purpose policy learning across tasks, datasets, and embodiments~\citep{oneill2024openx,ghosh2024octo,kim2024openvla}.
More recently, $\pi_0$, $\pi_{0.5}$, Hy-Embodied-0.5-VLA, and Xiaomi-Robotics-1 have expanded VLA capabilities in complex manipulation~\citep{black2024pi0,black2025pi05,zhang2026hyvla,xiaomi2026robotics1}.
These models use continuous action experts, large-scale heterogeneous data, and systematic training pipelines.
Collectively, these studies establish an effective approach: pre-trained VLMs represent scene and task semantics, while robot demonstrations teach mappings from multimodal observations to continuous actions.

\begin{figure}[H]
    \centering
    \includegraphics[width=\linewidth]{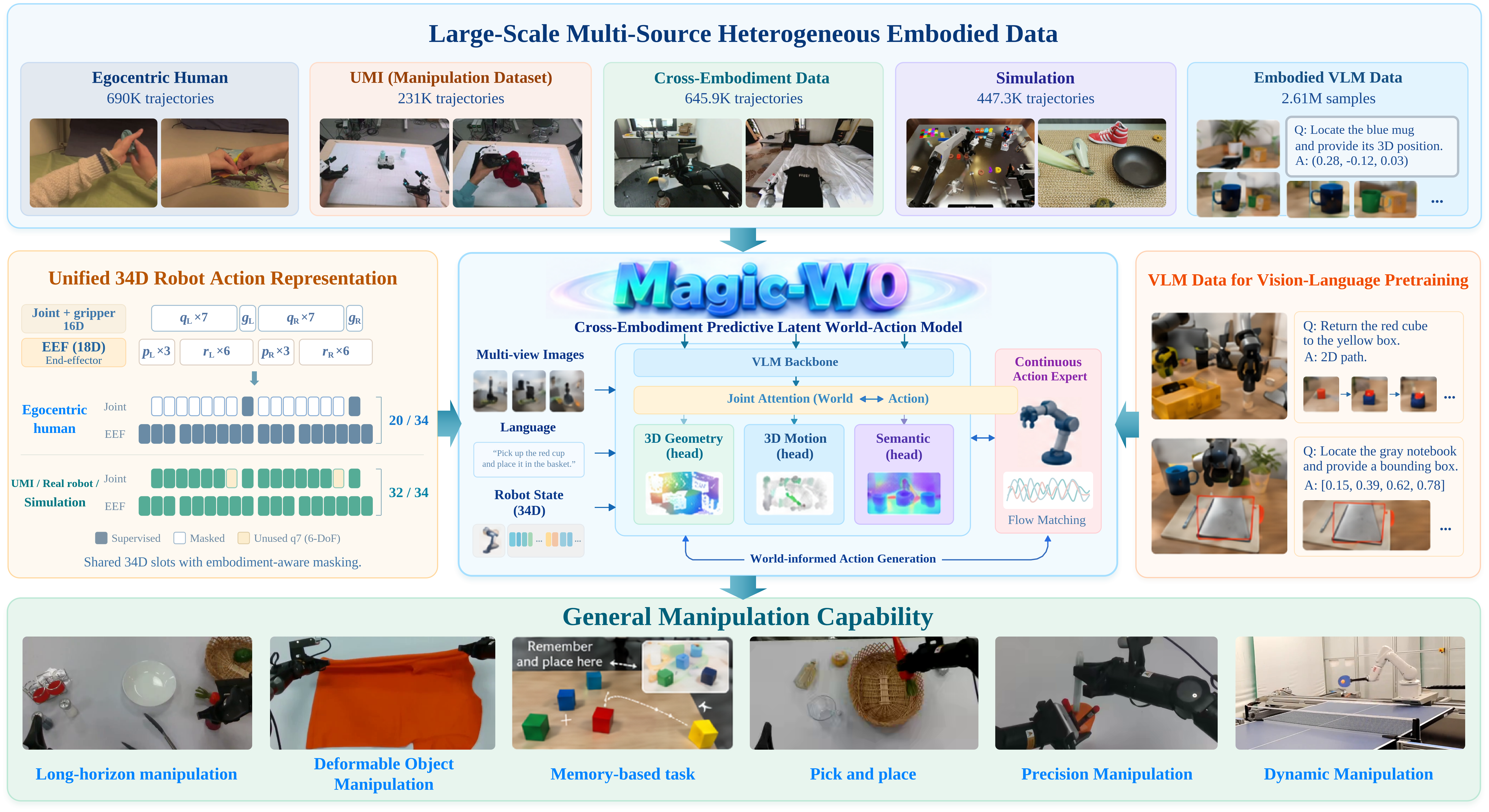}
    \caption{\textbf{Overview of Magic-W0, a large-scale, cross-embodiment structured world--action foundation model for physical intelligence.} The model jointly learns physical state transitions and continuous control from diverse embodied experience.}
    \label{fig:front-overview}
\end{figure}

Despite substantial progress, the supervision supporting these capabilities is asymmetric.
Vision--language pre-training provides rich priors on objects, scenes, and task semantics, while robot demonstrations directly constrain which actions a policy should execute under given observations.
By comparison, \emph{the state changes that candidate actions would induce in the current environment} are rarely constrained as explicitly through an independent prediction objective.
Existing VLAs can learn spatial relations, motion patterns, and physical interactions from robot trajectories.
However, they primarily acquire this knowledge implicitly while fitting actions.
Recent studies highlight different aspects of this distinction.
DreamZero notes that semantic generalization from vision--language priors does not automatically translate into generalization to unseen physical motions and interaction skills~\citep{ye2026dreamzero}.
Video Prediction Policy emphasizes that static visual representations alone cannot fully capture the temporal dynamics required for embodied tasks~\citep{hu2025vpp}.
Beyond task understanding and action generation, explicitly modeling \emph{action-conditioned environmental evolution} is therefore an important step toward stronger predictive modeling in robot policies.

\begin{figure}[!t]
    \centering
    \includegraphics[width=\linewidth]{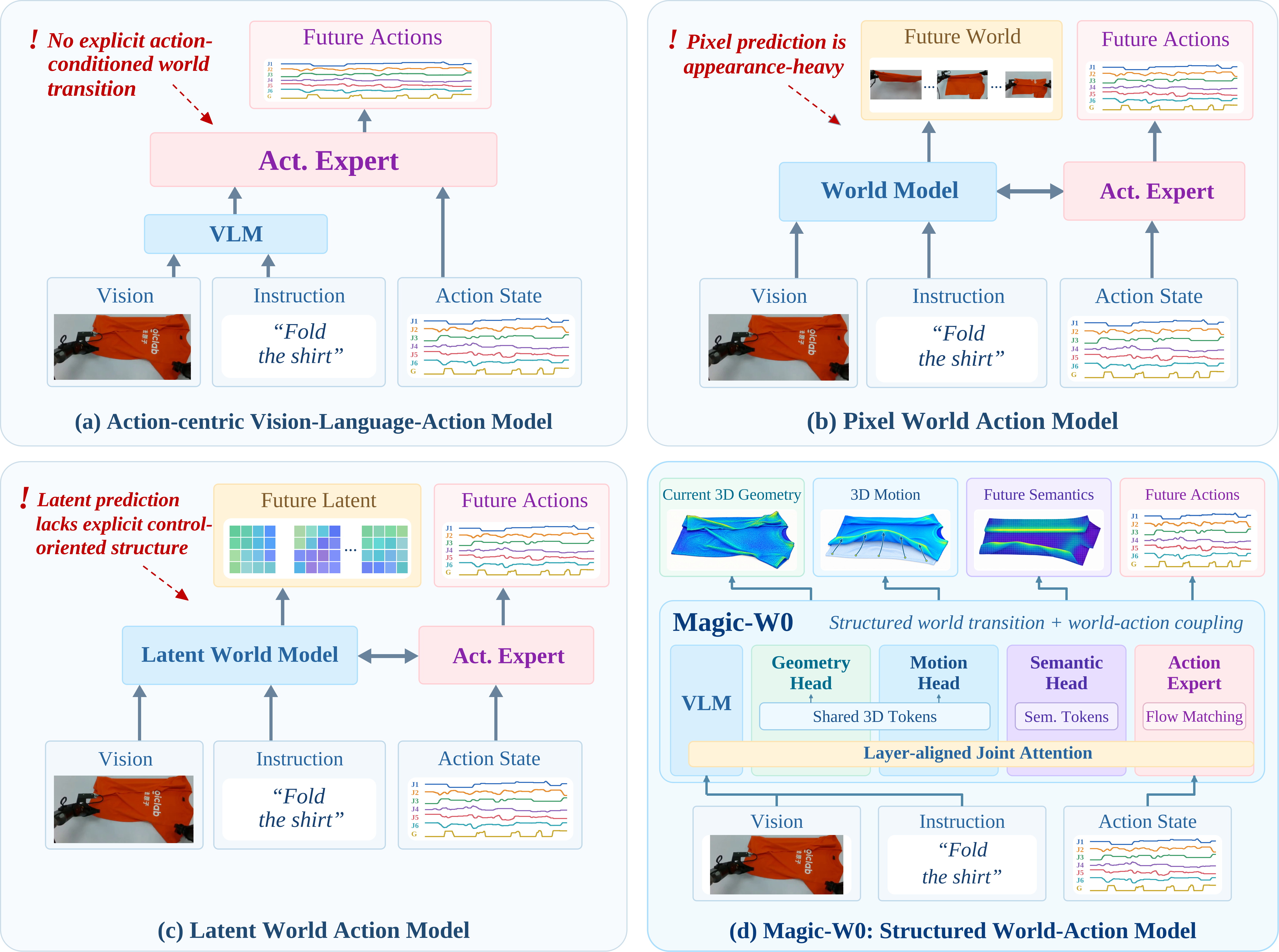}
    \caption{\textbf{Comparison of model architectures.} (a) Action-centered VLAs primarily rely on implicit world understanding. (b) Pixel-based WAMs predict future observations alongside actions. (c) Latent WAMs predict generic future representations alongside actions. (d) \model constructs Structured World Transition from Current 3D Geometry, 3D Motion, and Future Semantics, with bidirectional interaction through layer-aligned joint attention and an action expert.}
    \label{fig:intro-comparison}
\end{figure}

World modeling provides a natural way to introduce such predictive constraints.
World--action models (WAMs) jointly model current observations, actions, and future states, explicitly incorporating action-conditioned environmental evolution into policy learning.
DINO-WM treats predicting future outcomes from control actions as a key component of physical reasoning and planning~\citep{zhou2024dinowm}.
Fast-WAM further identifies action-conditioned future evolution modeling as a defining feature of WAMs relative to approaches focused solely on action generation~\citep{yuan2026fastwam}.
Unified World Models combine video and action modeling to incorporate environmental dynamics and interactions from videos without action annotations into policy pre-training~\citep{zhu2025uwm}.
V-JEPA~2 demonstrates the potential of large-scale video pre-training for learning action-conditioned latent world models~\citep{assran2025vjepa2}.
These studies extend policy learning from selecting actions under current conditions to predicting how the world will evolve under candidate actions.

As this direction develops, \emph{how to represent the future world} becomes a more fundamental question.
Existing methods broadly follow two approaches.
One predicts future images or videos in observation space, describing environmental evolution in an explicit, observable form~\citep{zhu2025uwm,ye2026dreamzero}.
Such representations retain rich scene information and support interpretable future rollouts.
However, complete future observations include texture, lighting, background, and viewpoint changes, many of which do not directly correspond to control-relevant state changes.
Moreover, appearance changes in two-dimensional observations do not explicitly reveal structure and motion in three-dimensional physical space.
The second approach predicts future visual features or intermediate representations in latent space, avoiding complete observation reconstruction.
DINO-WM directly predicts spatial features from a pre-trained vision model~\citep{zhou2024dinowm}.
Video Prediction Policy uses predictive visual representations within a video model to provide policies with future information~\citep{hu2025vpp}.
Fast-WAM shows that world modeling can benefit control without explicitly generating future observations at test time~\citep{yuan2026fastwam}.
Nevertheless, moving from observation space to latent space does not resolve how the future world should be structurally represented.
JEPA-WAM also identifies room for further design in the structure of predicted latent representations and their alignment with action representations~\citep{lin2026jepawam}.

For robot control, the current physical state, action-induced state transitions, and subsequent task-relevant outcomes serve distinct decision-making functions.
The current state defines the physical conditions for an action, the transition describes how a candidate action changes the environment, and the future state captures task-relevant outcomes.
Control-oriented world modeling therefore requires more than choosing between observation-space and latent prediction.
The key question is \textbf{how to organize predictive world representations with an explicit structure for decision-making}.
Alongside representation content, \textbf{how predictive world information participates in action generation} is central to world--action modeling.
Previous studies have explored latent dynamics-based planning~\citep{zhou2024dinowm,assran2025vjepa2}, policy conditioning on predictive features~\citep{hu2025vpp,yuan2026fastwam}, and world--action modeling within shared or joint generation frameworks~\citep{zhu2025uwm,cen2025worldvla,ye2026dreamzero,lin2026jepawam}.
These approaches progressively incorporate world prediction into policies, although predicting the future and using predictions to generate actions remain distinct problems.
A world prediction can serve as an auxiliary objective or additional condition without requiring evolving action hypotheses to continually inform future state prediction.
Likewise, predicted world changes need not continuously provide feedback across multiple stages of action representation.
This raises a second question: \textbf{how can action hypotheses and predicted world transitions mutually condition one another and evolve together during action generation within a unified policy?}

We propose \model, a structured world--action model for embodied robot manipulation, to address these two questions.
The model explicitly captures action-conditioned environmental evolution by jointly modeling structured predictive world representations and continuous action generation.
Figure~\ref{fig:intro-comparison} summarizes the differences between action-centered VLAs, generic WAMs, and our structured world--action model.
\model organizes control-relevant environmental evolution as \textbf{Structured World Transition}, using \emph{Current State--Transition--Future State} to describe the predictive process in robot decision-making.
Current State combines VLM context with Current 3D Geometry.
Transition uses 3D Motion to describe action-induced three-dimensional state transitions, while Future State uses Future Semantics to represent task-relevant future outcomes.
\model couples structured world prediction and continuous action generation in a unified modeling process.
Evolving action hypotheses participate in predicting future world transitions, while predicted world representations continuously inform action updates.
This jointly realizes \emph{action-conditioned world transition} and \emph{world-informed action generation}.
The design makes world prediction an internal predictive representation throughout continuous action generation, beyond an independent auxiliary objective or a conditioning feature applied only at the final stage.

\begin{samepage}
Our contributions are threefold:

\begin{itemize}
    \interlinepenalty=10000\relax
    \item \textbf{A structured world--action foundation model.} We propose Magic-W0, pre-trained on large-scale embodied data across embodiments.
    It models robot interaction as Current State--Transition--Future State.
    Current State combines VLM context with Current 3D Geometry, Transition uses 3D Motion to describe action-induced three-dimensional changes, and Future State represents task-relevant outcomes through Future Semantics.
    Together, these components form structured predictive world representations tailored to robot control.

    \item \textbf{Layer-aligned bidirectional coupling of world prediction and continuous action generation.} We design a world--action interaction architecture that continuously exchanges information between structured world representations and a continuous action expert at multiple network depths.
    Action hypotheses inform future world transitions, while predicted world representations provide continuous feedback to action updates.
    This supports mutual conditioning and joint evolution through action-conditioned world transition and world-informed action generation.

    \item \textbf{A unified cross-embodiment representation for heterogeneous data from multiple sources.} Egocentric human manipulation, UMI, real-robot, and simulation data differ in embodiment structure, action space, and supervision.
    We construct a unified 34-dimensional state--action interface that maps available end-effector poses, gripper information, and joint states to corresponding dimensions.
    Unified coordinate conventions and action representations enable world--action pre-training across these sources.
\end{itemize}
\end{samepage}

\section{Related Work}
\label{sec:related-work}

\paragraph{General-purpose VLA policies and action experts.}
General-purpose vision--language--action policies have evolved around translating semantic knowledge from vision--language pre-training into scalable robot control.
RT-1 and RT-2 first demonstrated the value of large-scale robot data and internet semantic knowledge for task coverage and instruction generalization~\citep{brohan2022rt1,zitkovich2023rt2}.
Open X-Embodiment, Octo, and OpenVLA subsequently extended this paradigm to training across datasets, tasks, and embodiments~\citep{oneill2024openx,ghosh2024octo,kim2024openvla}.
For action modeling, Diffusion Policy represents continuous control as a conditional diffusion process, while RDT-1B scales diffusion-based action generation to large-scale bimanual manipulation~\citep{chi2023diffusionpolicy,liu2024rdt1b}.
FAST revisits action discretization in autoregressive VLAs through high-frequency action compression~\citep{pertsch2025fast}.
$\pi_0$, $\pi_{0.5}$, Hy-Embodied-0.5-VLA, and Xiaomi-Robotics-1 further improve complex manipulation using continuous action experts, heterogeneous data, and systematic training pipelines~\citep{black2024pi0,black2025pi05,zhang2026hyvla,xiaomi2026robotics1}.
Being-H0.5 pre-trains on human interaction data and supports cross-embodiment knowledge transfer through a unified action space and mixture-of-flow action experts~\citep{luo2026beingh05}.
GR-2 introduces large-scale video priors into generative VLAs, while GR00T N1, Gemini Robotics, and G0 explore dual-system control or embodied reasoning~\citep{cheang2024gr2,nvidia2025gr00tn1,gemini2025robotics,jiang2025g0}.
X-VLA uses soft prompts to accommodate heterogeneity across embodiments~\citep{zheng2025xvla}.
These advances leave open how to jointly organize current three-dimensional geometry, three-dimensional motion, and future semantics as explicit predictive representations within a policy.
\model retains the effective combination of a VLM backbone and a continuous action expert.
Its focus is to make three-dimensional states and future changes explicit predictive structures within the policy, grounding actions in world transition modeling.

\paragraph{World--action models: from explicit generation to predictive latent representations.}
The representation used to describe the future determines which control-relevant environmental changes a WAM retains.
Explicit approaches model observable futures.
UniPi, Dreamitate, and RoboDreamer convert language or task conditions into future frames or visual plans~\citep{du2023unipi,liang2024dreamitate,zhou2024robodreamer}.
Cosmos Policy, LingBot-VA, and DiT4DiT learn environmental evolution using video priors, shared latent spaces, or cascaded diffusion~\citep{kim2026cosmospolicy,li2026causalworld,ma2026dit4dit}.
These representations are intuitive and support visual rollouts.
However, complete future observation reconstruction allocates capacity to texture, lighting, and background details that are weakly related to control.
Two-dimensional image changes also cannot fully describe the three-dimensional physical space surrounding a robot.
To reduce reconstruction demands, Video Prediction Policy, Unified Video Action Model, and Fast-WAM use intermediate predictive features or future latent representations that require no decoding~\citep{hu2025vpp,li2025uva,yuan2026fastwam}.
Being-H0.7 introduces learnable latent queries between perception and action.
It aligns a prior branch driven by current observations with a posterior branch driven by future observations to learn predictive representations for action generation~\citep{luo2026beingh07}.
Only the prior branch is retained at inference.
Being-H0.8 additionally includes future visual and tactile information in posterior supervision, incorporating contact-related interaction information into latent world states~\citep{beingbeyond2026beingh08}.
These studies shift attention from whether to predict the future to how to organize latent world representations for control.
\model jointly learns Current 3D Geometry, 3D Motion, and Future Semantics in latent space.
The respective targets are current-frame geometry, cross-time three-dimensional motion, and future-frame semantic features, representing the current scene structure and its subsequent changes.

\paragraph{Three-dimensional geometry and dynamic scene representations.}
Three-dimensional representations connect image content to the physical space in which robots act.
Policy-based approaches have begun incorporating three-dimensional knowledge into VLAs.
SpatialVLA explicitly encodes three-dimensional positions, while Spatial Forcing improves spatial understanding by aligning intermediate features from a three-dimensional foundation model without additional depth inputs~\citep{qu2025spatialvla,li2025spatialforcing}.
For world modeling, TesserAct, X-WAM, SpatialVAM, and RynnWorld-4D explicitly predict three-dimensional structure and its temporal evolution~\citep{zhen2025tesseract,guo2026xwam,li2026mvdp,zhao2026rynnworld4d}.
Their respective representations are RGB-DN, multi-view RGB-D, multi-view heatmap and RGB videos, and RGB-DF.
These studies show that geometry and dynamics can improve action learning, primarily through spatial representation alignment or observable 4D world reconstruction.
Complementary general-purpose vision models provide transferable representations for latent supervision.
Depth Anything 3 recovers a unified three-dimensional visual space from arbitrary views, while DINOv3 learns high-quality dense features that preserve object, region, and scene semantics~\citep{lin2025da3,simeoni2025dinov3}.
Track4World recovers current geometry using a DA3 backbone adapted through training on dynamic videos.
Its 3D motion head estimates dense three-dimensional motion in a unified world coordinate system~\citep{lu2026track4world}.
\model uses the geometry and motion representations at different levels of Track4World, together with future DINOv3 semantic features, as three mutually constraining learning objectives.
Current geometry anchors the state, future motion describes the three-dimensional transition, and future semantics captures the significance of that transition for the task.
The model learns these latent representations rather than reproducing the teachers' observable outputs.

\paragraph{Coupling world modeling with action generation.}
The use of world prediction in control has progressed from latent planning to joint modeling within policies.
Dreamer optimizes policies through latent imagination~\citep{hafner2020dreamer}.
TD-MPC2 combines latent dynamics, value estimation, and local trajectory optimization within a decoder-free implicit world model~\citep{hansen2024tdmpc2}.
Together, these methods establish a basic paradigm for predictive models supporting continuous control.
For robot tasks, DINO-WM and V-JEPA~2 use action-conditioned visual latent representations for planning~\citep{zhou2024dinowm,assran2025vjepa2}.
Video Prediction Policy and Fast-WAM directly incorporate predictive features into policy conditioning~\citep{hu2025vpp,yuan2026fastwam}.
More tightly coupled approaches jointly learn world changes and actions within shared networks.
GR-1 and GR-2 predict future images and robot actions together~\citep{wu2023gr1,cheang2024gr2}.
Unified World Models, WorldVLA, DreamZero, and JEPA-WAM connect the two objectives through coupled diffusion, autoregressive generation, video diffusion backbones, and joint embedding prediction, respectively~\citep{zhu2025uwm,cen2025worldvla,ye2026dreamzero,lin2026jepawam}.
Being-M0.7 jointly predicts future visual latent representations and whole-body motion representations.
It conditions an action expert on intermediate features from the predictive prior, using human video and motion priors for humanoid mobile manipulation~\citep{yue2026beingm07}.
Being-H0.8 reuses world-state context while integrating updated proprioceptive states and tactile feedback through slow--fast action experts to dynamically correct imminent action segments~\citep{beingbeyond2026beingh08}.
Through latent planning, predictive feature conditioning, or shared generation, these studies progressively make world prediction part of action learning.
\model further examines representation content and interaction depth within this coupling.
It constructs Structured World Transition from Current 3D Geometry, 3D Motion, and Future Semantics.
A layer-aligned architecture with multiple expert streams mutually conditions action hypotheses and structured world transitions at multiple network depths.
This jointly supports action-conditioned world transition and world-informed action generation.

\section{Magic-W0}
\label{sec:method}

\begin{figure}[!htbp]
    \centering
    \includegraphics[width=\linewidth]{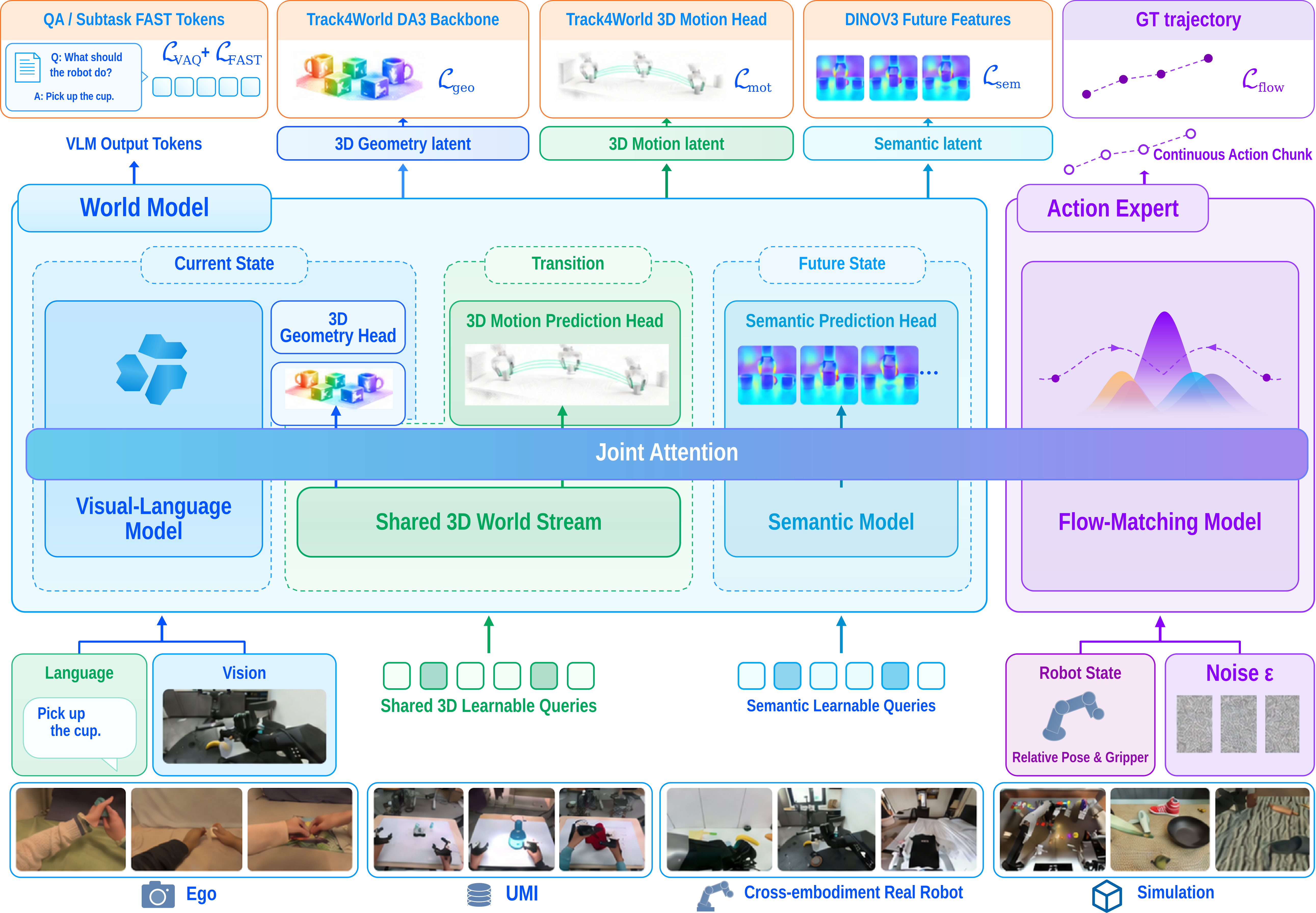}
    \caption{\textbf{Overall framework of \model.} Structured World Transition comprises Current State, Transition, and Future State. Current State combines vision--language context with Current 3D Geometry; Transition is represented by 3D Motion; Future State is represented by Future Semantics. Current 3D Geometry and 3D Motion share the hidden states of the 3D stream, while a separate semantic stream models Future Semantics. Both streams interact with the action expert through layer-aligned joint attention, with vision--language context supplied by the VLM. Track4World and DINOv3 provide latent supervision only during training.}
    \label{fig:overview}
\end{figure}

\model combines structured world prediction and continuous action generation within a single model.
Given multi-view observations $\obs_{\leq t}$, a language instruction $\lang$, and a proprioceptive state $\state_t$, it generates an action chunk $\action_{t:t+H-1}$ of length $H$.
It also models Current 3D Geometry, 3D Motion, and Future Semantics in latent space.
Structured World Transition uses two world representation branches: a 3D stream jointly models Current 3D Geometry and 3D Motion, while a separate semantic stream models Future Semantics.
As shown in Figure~\ref{fig:overview}, the VLM backbone provides task context.
Layer-aligned interaction among the 3D stream, semantic stream, and action expert mutually conditions structured world representations and continuous action generation within the same computation.

\Needspace{9\baselineskip}
\subsection{Structured World Representations}
\label{sec:structured-world-representations}

To describe world-state changes during robot interaction, we organize structured world representations into three stages: Current State--Transition--Future State.
As shown in Figure~\ref{fig:overview}, Current State includes VLM context extracted from current observations, language instructions, and proprioceptive states, together with Current 3D Geometry predicted by the 3D stream.
VLM context provides scene and task semantics, while Current 3D Geometry describes the spatial structure before action execution.
3D Motion describes action-induced three-dimensional state transitions, and Future Semantics captures the task-relevant scene state after the transition.
Let $\mathbf C_t$ denote VLM context. Structured World Transition is written as
\begin{equation}
\mathcal{W}_{t,\Delta}=\left(
\underbrace{\mathbf C_t,\mathbf Z_t}_{\mathrm{Current\ State}},\;
\underbrace{\mathbf Z_{t\rightarrow t+\Delta}}_{\mathrm{Transition}},\;
\underbrace{\mathbf Z_{t+\Delta}}_{\mathrm{Future\ State}}
\right).
\label{eq:structured-world-transition}
\end{equation}
Here, $\mathbf Z_t$, $\mathbf Z_{t\rightarrow t+\Delta}$, and $\mathbf Z_{t+\Delta}$ denote Current 3D Geometry, 3D Motion, and Future Semantics, respectively.
All three are defined directly in their corresponding teacher feature spaces.
Current State is jointly represented by $\mathbf C_t$ and $\mathbf Z_t$, while the motion and semantic representations correspond to the same future time.
$t$ denotes observation time, and $\Delta$ aligns the future time with the end of the action chunk's coverage interval.
Its value depends on the source-specific action sampling multiplier (Section~\ref{sec:temporal-alignment}).

\paragraph{Teacher representation targets.}
Geometry and motion supervision comes from a frozen Track4World model~\citep{lu2026track4world}.
The teacher processes the current--future frame pair in a single forward pass.
Its DA3 backbone, trained on dynamic scenes, provides source-frame geometry features~\citep{lin2025da3}, while its 3D motion head provides cross-time motion features.
Geometry targets are extracted from the current frame and remain fixed for each sample, independent of candidate actions.
Future semantic targets use patch features from a frozen DINOv3 model applied to future frames, supervising task outcomes through latent semantic features~\citep{simeoni2025dinov3}.

All three targets are represented as $16\times16\times1024$ spatial features, supervising each prediction head on its corresponding query grid.
Teachers extract latent targets only during training; the internal 3D and semantic streams generate representations at inference.

\subsection{World--Action Co-Modeling}
\label{sec:world-action-comodeling}

\begin{figure}[!htbp]
    \centering
    \includegraphics[width=\linewidth]{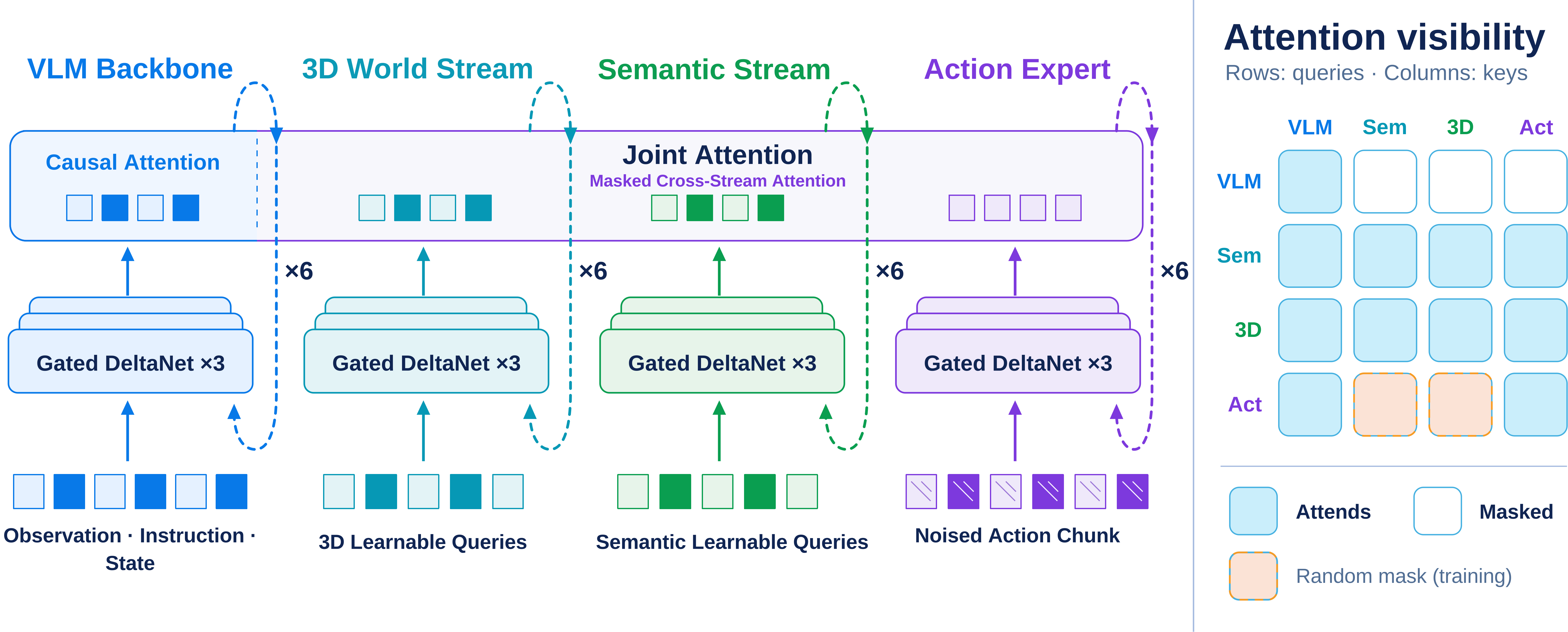}
    \caption{\textbf{Layer-aligned world--action interaction.} Each block contains three within-stream Gated DeltaNet layers and one attention layer. Six repetitions yield 24 aligned depths. The VLM retains causal computation at attention layers, while the semantic stream, 3D stream, and action expert access joint keys and values through joint attention. In the visibility matrix, rows indicate query sources and columns indicate key--value sources. Blue denotes visible connections; white denotes masked connections. Orange dashed cells in the Act row and Sem/3D columns indicate connections that can be randomly masked during training. Only one is selected per masking event; semantic and 3D query visibility remains unchanged.}
    \label{fig:architecture}
\end{figure}

The 3D stream, semantic stream, and action expert are computed within the same forward pass.
World predictions can therefore respond to action inputs and condition action updates.
Figure~\ref{fig:architecture} details this interaction.
The figure labels ``VLM Backbone,'' ``Semantic Stream,'' ``3D World Stream,'' and ``Action Expert'' correspond to the VLM backbone, semantic stream, 3D stream, and action stream, respectively.
The four components are aligned in depth, alternating within-stream computation and cross-stream information exchange.

\subsubsection{Inputs and Bidirectional Interaction}
\label{sec:world-action-inputs}
Qwen3.5-2B encodes multi-view visual tokens, language tokens, and a projected proprioceptive state token into task context $\mathbf C_t$.
Action space type and joint dimensionality enter the prefix as textual metadata.
The action expert generates continuous actions through flow matching~\citep{lipman2023flowmatching}, representing a normalized action chunk as $H$ tokens.
Let $\tau\in[0,1]$ denote flow time, independent of observation time $t$, and let $\mathbf x_\tau$ denote the noisy action chunk at that time.
The action stream combines a projection of noisy actions, within-chunk positional embeddings, flow time, and proprioceptive state conditioning.
It outputs the velocity field $\mathbf v_\theta(\mathbf x_\tau,\tau)$.

\paragraph{Bidirectional interaction.}
The VLM provides task context to the three expert streams while updating independently along its native causal path.
The 3D, semantic, and action streams access one another's intermediate hidden states through joint attention.
Semantic and 3D tokens attend to action tokens, allowing current action hypotheses to inform future motion and semantic predictions.
This realizes \emph{action-conditioned world transition}.
Action tokens attend to both world streams, using spatial structure, dynamic changes, and task semantics to update the velocity field.
This realizes \emph{world-informed action generation}.
The geometry and motion heads read representations from the final 3D hidden states, while the semantic head reads from the final semantic hidden states.
These outputs receive supervision in teacher feature spaces.
Action generation uses the world hidden states updated at successive layers.
Geometry targets remain fixed, although shared 3D hidden states change with action inputs.

\subsubsection{Layer-Aligned Joint Attention}
\label{sec:layer-aligned-attention}
The semantic, 3D, and action streams have independent parameters and remain aligned in depth with the VLM.
The network has 24 layers, with one attention layer after every three Gated DeltaNet layers~\citep{yang2025gdn}.
Information is exchanged at layers 4, 8, \ldots, 24.
Gated DeltaNet performs within-stream updates.
At joint attention layers, each expert stream uses its own queries to attend to joint keys and values:
\begin{equation}
\begin{aligned}
(\mathbf K_{\mathrm{joint}}^\ell,\mathbf V_{\mathrm{joint}}^\ell)
&=\operatorname{Concat}\!\left[
(\mathbf K,\mathbf V)_{\mathrm{vlm}}^\ell,
(\mathbf K,\mathbf V)_{\mathrm{sem}}^\ell,
(\mathbf K,\mathbf V)_{\mathrm{3d}}^\ell,
(\mathbf K,\mathbf V)_{\mathrm{act}}^\ell\right],\\
\mathbf Z_m^{\ell+1}
&=\mathcal{F}_m^\ell\!\left(\mathbf Q_m^\ell,\mathbf K_{\mathrm{joint}}^\ell,\mathbf V_{\mathrm{joint}}^\ell\right),
\quad m\in\{\mathrm{sem},\mathrm{3d},\mathrm{act}\}.
\end{aligned}
\label{eq:joint-stream-update}
\end{equation}
Here, $\ell$ is the layer index, and $\mathcal{F}_m^\ell$ is the update operation for expert stream $m$.
The VLM computes prefix keys and values through its own attention projections at layer $\ell$; the expert streams directly reuse them.
Each stream reads the hidden states before the update, and all streams synchronously produce their next-layer representations.
The 3D and semantic streams share a starting position after the prefix.
Action token positions follow the longer of these two sequences, placing all three streams in a common positional coordinate system.

Each query stream accesses keys and values according to the visibility matrix in Figure~\ref{fig:architecture}.
The VLM does not attend to expert streams, while the three expert streams can access one another by default.
Section~\ref{sec:pretrain-schedule} describes random masking of action-query access to either the 3D or semantic stream during training.
Joint computation continues at every action integration step during inference.
Section~\ref{sec:inference-execution} details caching, integration, and action execution.

\subsection{Training Objectives}
\label{sec:objectives}

\paragraph{World representation supervision.}
Each world prediction is aligned with its corresponding teacher target.
Let $\hat{\mathbf z}_i^r$ and $\mathbf y_i^r$ denote the student prediction and teacher feature for spatial token $i$, where $r\in\{\mathrm{geo},\mathrm{mot},\mathrm{sem}\}$.
Semantic predictions use cosine distance to align feature directions.
Geometry and motion predictions use mean squared error after channel normalization:
\begin{equation}
\begin{aligned}
\ell_{\mathrm{sem}}(\hat{\mathbf z},\mathbf y)&=1-\cos(\hat{\mathbf z},\mathbf y),\\
\ell_r(\hat{\mathbf z},\mathbf y)&=\frac{1}{d}\|\operatorname{LN}(\hat{\mathbf z})-\operatorname{LN}(\mathbf y)\|_2^2,
\quad r\in\{\mathrm{geo},\mathrm{mot}\}.
\end{aligned}
\label{eq:world-token-losses}
\end{equation}
Here, $d=1024$, and $\operatorname{LN}$ denotes parameter-free channel normalization.
Each $\Loss_r$ averages $\ell_r$ over valid tokens, masking out missing views and invalid teacher outputs.
Geometry and motion losses jointly constrain shared 3D hidden states, while the semantic loss constrains future predictions from the separate semantic stream.

\paragraph{Continuous action loss.}
For a normalized action chunk $\action$ and Gaussian noise $\bm{\epsilon}\sim\mathcal{N}(\mathbf 0,\mathbf I)$, we construct $\mathbf x_\tau=(1-\tau)\action+\tau\bm{\epsilon}$.
The model regresses the target velocity $\mathbf u^\star=\bm{\epsilon}-\action$:
\begin{equation}
\Loss_{\mathrm{flow}}=
\mathbb{E}_{\action,\bm{\epsilon},\tau}\!\left[
\|\mathbf v_\theta(\mathbf x_\tau,\tau)-\mathbf u^\star\|_{\mathbf M_{\mathrm{act}}}^2\right].
\label{eq:flow-objective}
\end{equation}
Here, $\|\cdot\|_{\mathbf M_{\mathrm{act}}}^2$ denotes mean squared error over valid action timesteps and dimensions.
Appendix~\ref{app:losses} provides the flow-time sampling distribution and full mask normalization for all losses.

\paragraph{Joint objective.}
The three world representation losses are optimized jointly with the continuous action loss.
The total training loss is
\begin{equation}
\begin{aligned}
\Loss={}&
\lambda_{\mathrm{flow}}\Loss_{\mathrm{flow}}
+\lambda_{\mathrm{VQA}}\Loss_{\mathrm{VQA}}
+\lambda_{\mathrm{FAST}}\Loss_{\mathrm{FAST}}
+\lambda_{\mathrm{sem}}\Loss_{\mathrm{sem}}\\
&+\lambda_{\mathrm{3d}}\left(
\lambda_{\mathrm{geo}}\Loss_{\mathrm{geo}}
+\lambda_{\mathrm{mot}}\Loss_{\mathrm{mot}}\right).
\end{aligned}
\label{eq:total-objective}
\end{equation}
$\Loss_{\mathrm{FAST}}$ provides discrete action supervision during pre-training.
$\Loss_{\mathrm{VQA}}$ is autoregressive cross-entropy for visual question answering (VQA), computed only when VQA supervision is enabled and valid question--answer annotations are available.
Appendix~\ref{app:losses} defines these losses and their masks.
Vision--language batches use standard VLM autoregressive cross-entropy.
Section~\ref{sec:pretrain-schedule} describes batch mixing, gradient paths, and auxiliary weight schedules.

\section{Pre-Training}
\label{sec:data}
\label{sec:pretraining}

\model is pre-trained jointly on manipulation trajectories from multiple sources and embodied vision--language data.
Manipulation trajectories supervise continuous action generation and structured world representations, while vision--language data maintain the backbone's capacity to model scenes, spatial relations, and task semantics.
We first map heterogeneous sources to a unified state--action interface.
We then construct future supervision aligned with the action chunk's temporal coverage and jointly optimize world prediction, action generation, and vision--language objectives.

\subsection{Pre-Training Corpus and Unified Representation}
\label{sec:data-composition}
\label{sec:cross-embodiment}

\begin{figure}[tbp]
\centering
\includegraphics[width=0.9\linewidth]{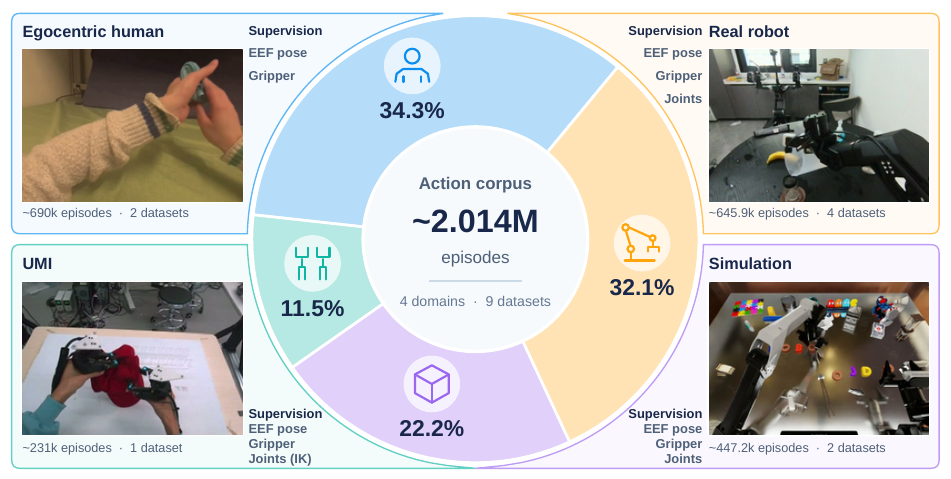}
\caption{\textbf{Approximate numbers of valid manipulation episodes used for pre-training, aggregated by acquisition domain.} Appendix Table~\ref{tab:embodied-pretraining-corpus} reports dataset-level counts. Both individual counts and the total are approximate.}
\label{fig:action-corpus-distribution}
\end{figure}

\begin{figure}[htbp]
\centering
\includegraphics[width=\linewidth]{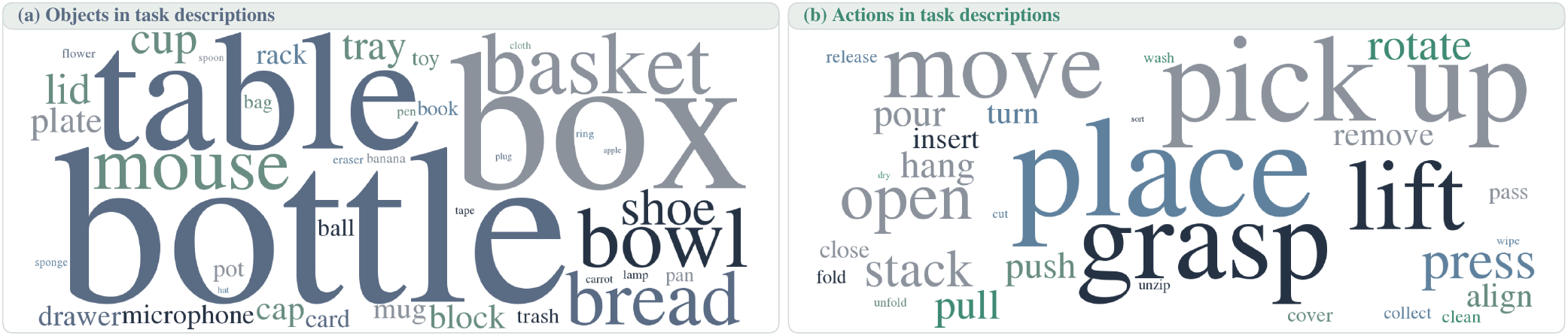}
\caption{\textbf{Example object and action words in trajectories with task descriptions.} Font size reflects mention frequency in the text.}
\label{fig:task-vocabulary}
\end{figure}

\paragraph{Embodied pre-training corpus.}
The manipulation corpus used for pre-training spans four acquisition domains: egocentric human manipulation, UMI, real robots, and simulation.
It comprises approximately 2.014 million valid episodes after filtering and aggregation (Figure~\ref{fig:action-corpus-distribution}).
Egocentric human manipulation data provide end-effector pose and gripper supervision.
UMI, real-robot, and simulation data additionally provide joint-space supervision; UMI joint targets are constructed through kinematic retargeting.
Appendix Table~\ref{tab:embodied-pretraining-corpus} lists the approximate number of valid episodes for each dataset.
The vision--language supervised fine-tuning corpus includes EO-Data1.5M, Robo2VLM-1, and in-house annotations constructed from selected open-source datasets.
It totals approximately 2.61 million samples, with sources and sizes listed in Appendix Table~\ref{tab:vlm-sft-corpus}.
These samples supervise spatial localization, task semantics, and interleaved image--text--action modeling.
Manipulation and vision--language batches are mixed at a $9:1$ ratio; the latter train the backbone through an autoregressive objective.
Some manipulation trajectories contain action descriptions and scene labels.
Figure~\ref{fig:task-vocabulary} shows example object and action words in task descriptions.

\begin{figure}[htbp]
\centering
\includegraphics[width=\linewidth]{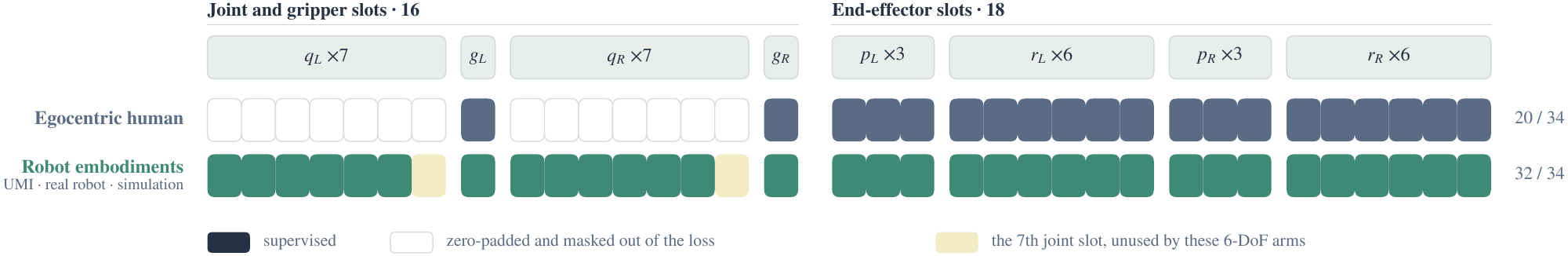}
\caption{\textbf{Slot occupancy in the 34-dimensional state--action interface.} $q$ denotes joint angles, $g$ gripper opening, $p$ end-effector position, and $r$ the 6D rotation encoding. Subscripts $L$ and $R$ distinguish sides; blank slots are zero-padded and masked. Egocentric human data have no corresponding robot embodiment, so all 14 joint slots are invalid. The example UMI, real-robot, and simulation trajectories share the same occupancy and are grouped in one row. Each uses two arms with six degrees of freedom, leaving the seventh joint slot of each arm empty.}
\label{fig:slot-schema}
\end{figure}

\paragraph{Unified state--action representation.}
\label{sec:unified-state-action-interface}
Data sources differ substantially in joint counts, control interfaces, reference frames, and action scales.
We map all manipulation data to a 34-dimensional state--action interface with fixed semantics:
\begin{equation}
\mathbf{x}=\left[
\mathbf{q}_L,g_L,\mathbf{q}_R,g_R,
\mathbf{p}_L^{C_t},\mathbf{r}_L^{C_t},
\mathbf{p}_R^{C_t},\mathbf{r}_R^{C_t}\right]\in\mathbb{R}^{34},
\label{eq:unified-action-schema}
\end{equation}
where $L$ and $R$ denote the left and right sides.
Each side contains seven joint dimensions $\mathbf q$, one gripper opening dimension $g$, three end-effector position dimensions $\mathbf p$, and six rotation dimensions $\mathbf r$.
The superscript $C_t$ denotes the camera coordinate frame at time $t$.
Each source populates its observable slots; the remaining dimensions are zero-padded and excluded from training losses through validity masks.
Figure~\ref{fig:slot-schema} illustrates slot occupancy.
New embodiments populate the corresponding slots without changing model or checkpoint parameter shapes.

Real-robot and simulation data map joint and end-effector states according to source metadata.
UMI uses kinematic retargeting to convert handheld-device trajectories into joint, gripper, and end-effector supervision for a target robot.
Egocentric human data use hand motion to construct virtual end-effector poses and continuous gripper openings; joint dimensions remain invalid.
All sources use unified physical units, camera-frame end-effector poses, 6D rotation encoding, and source-specific action scale normalization.
Appendix~\ref{app:data} details coordinate transformations, retargeting, and training sample construction.

\subsection{Action-Aligned Training Samples}
\label{sec:temporal-alignment}
\label{sec:supervision-construction}

Action sampling multipliers differ across sources, so action chunks of equal length cover different intervals in the original observation sequences.
Both 3D Motion and Future Semantics require future observations to supervise Structured World Transition.
Using the same future-frame offset for all sources could misalign the selected future state with the interaction interval actually covered by the action chunk.
We therefore introduce source-aware action--world temporal alignment.
Future observations are selected according to each source's action timescale, so continuous action and future world supervision describe the same interaction from the same current state.

Each training sample is anchored at time $t$.
It contains multi-view observations $\obs_{\leq t}$, a language instruction $\lang$, a proprioceptive state $\state_t$, an action chunk $\action_{t:t+H-1}$, and the temporally corresponding future observation $\obs_{t+\Delta_i}$.
We use $H=50$ throughout this stage.
Let $\rho_i$ denote the action sampling multiplier for source $i$.
The future-frame offset corresponding to a chunk of length $H$ on the original observation timeline is
\begin{equation}
\Delta_i=\left\lceil\frac{H}{\rho_i}\right\rceil,
\label{eq:source-aware-horizon}
\end{equation}
aligning the future observation with the end of the action chunk's coverage interval.
Egocentric human manipulation typically involves faster motion and uses action sampling multipliers greater than one.
For example, EgoDex uses $\rho_i=1.95$.
With $H=50$, $H/\rho_i\approx25.64$ frames, yielding $\Delta_i=26$ after rounding up according to Equation~\eqref{eq:source-aware-horizon}.
Teleoperated data typically use $\rho_i$ between $0.9$ and $1.2$, with future-frame offsets determined separately for each source.
Despite differences in action temporal resolution, $\action_{t:t+H-1}$ and $\obs_{t+\Delta_i}$ correspond to the same state evolution interval within each source.

After temporal alignment, current and future observations jointly construct the three types of Structured World Transition supervision.
Current 3D Geometry uses the head-camera view at time $t$ to represent the current structure before action execution.
3D Motion uses the head-camera frame pair at $t$ and $t+\Delta_i$ to describe the three-dimensional state transition over the action chunk.
Future Semantics uses available multi-view observations at $t+\Delta_i$ to represent future scene semantics after the transition.
These targets correspond to Current State, Transition, and Future State, sharing an anchor time and a future time.
Three-dimensional supervision uses only the head-camera view; semantic supervision uses all available views.
Frozen teacher models extract world targets online during training forward passes, and each branch learns latent representations directly in its corresponding teacher feature space.
Continuous actions use a chunk-wise delta representation: every timestep's control target is expressed relative to the chunk's initial state.
Joint angles and end-effector positions use relative changes, rotations use relative rotations, and grippers retain absolute openings.

\subsection{Joint Pre-Training}
\label{sec:pretrain-schedule}
After unifying cross-source state--action representations and aligning action--world timing, pre-training must coordinate continuous control, structured world modeling, and vision--language representation learning.
Magic-W0 uses joint pre-training with multiple objectives to learn action generation and Structured World Transition within the same model.
It also preserves the backbone's representation of scenes, spatial relations, and task semantics.

Continuous action generation is supervised through flow matching.
Current 3D Geometry, 3D Motion, and Future Semantics respectively supervise current physical structure, action-induced state transitions, and future semantic representations.
A FAST discrete action objective additionally provides explicit robot action supervision to the VLM backbone.
Vision--language modeling uses a standard autoregressive objective to preserve the pre-trained backbone's multimodal understanding.
Controlled gradient paths and auxiliary objective weights coordinate these signals to reduce interference between objectives.
The following paragraphs describe gradient isolation, auxiliary weight schedules, and cross-stream visibility.
Appendix~\ref{app:losses} defines the losses and mask normalization.

\paragraph{Gradient isolation.}
Pre-training uses gradient isolation, or knowledge insulation~\citep{driess2025ki}.
The 3D stream, semantic stream, and action expert can access VLM context through joint attention.
However, continuous action and world representation losses do not backpropagate through this context or its key--value projections.
Only the vision--language autoregressive objective and FAST discrete action supervision update the VLM backbone; action and world objectives optimize the expert streams.
The visual encoder remains trainable.

\paragraph{Auxiliary objectives and stream visibility.}
FAST, semantic, and three-dimensional supervision receive relatively high auxiliary weights early in training, followed by gradual decay.
This increases the relative emphasis on world supervision early on and reduces the influence of auxiliary objectives later.
Appendix~\ref{app:pretrain-config} provides the weights, schedules, and optimization settings.

To reduce the action expert's reliance on a single world branch, we randomly mask its access to one branch with a small probability per robot batch.
Each masking event selects either the 3D or semantic stream; visibility and supervision for the 3D and semantic queries remain unchanged.
In most batches, action queries access both world streams, learning to use three-dimensional structure and future semantics together.
These masked batches also expose the model to conditions matching inference-time masking of either source.

\section{Post-Training and Inference}
\label{sec:training-inference}

After pre-training on large-scale embodied data from multiple sources, \model adapts to deployment scenarios through supervised fine-tuning in the target domain.
Post-training retains Structured World Transition and the world--action interaction architecture, transferring structured world modeling and action generation together to downstream tasks.
At inference, the 3D stream, semantic stream, and action expert are jointly updated throughout flow-based generation.
This carries action-conditioned world transition and world-informed action generation into robot control.

\subsection{Supervised Fine-Tuning in the Target Domain}
\label{sec:finetuning}

For each target domain, we initialize the model from the full pre-trained checkpoint and jointly fine-tune on all demonstrations in that domain.
This produces a unified policy covering multiple tasks within the domain.
Fine-tuning preserves model topology and world representations.
The unified state--action interface, current--future temporal alignment, and latent world supervision follow the pre-training settings.
The 3D stream, semantic stream, and action expert continue to interact within a single forward pass.
Downstream adaptation thus jointly adjusts pre-trained world representations and action generation using target-domain data, rather than training a separate action policy from scratch.

\paragraph{Post-training objectives.}
Fine-tuning uses the continuous action and world representation losses defined in Section~\ref{sec:objectives}.
It excludes FAST discrete action supervision and vision--language corpus mixing.
The gradient isolation used during pre-training is removed.
Continuous action and world prediction objectives jointly update the VLM backbone, 3D stream, semantic stream, and action expert; the visual encoder also remains trainable.
World representation losses use fixed weights during fine-tuning.

\paragraph{Target-domain configuration.}
Target-domain data use the unified state--action interface, with validity masks selecting dimensions that are observable and controllable.
For embodiments with only end-effector control, supervision covers end-effector position, rotation, and gripper slots, while unused joint dimensions are masked out.
Observation views for world supervision depend on the target domain's sensor configuration.
When future times extend beyond a trajectory's endpoint, we hold the endpoint values to construct action and world supervision, retaining terminal training samples.

\subsection{Inference-Time Execution}
\label{sec:inference-execution}

At inference, \model first encodes current multi-view observations, language instructions, and proprioceptive states into vision--language context.
It caches the VLM's keys and values at each layer.
An action chunk of length $H$ is initialized from Gaussian noise, and continuous actions are generated by numerically integrating the flow-matching velocity field.
The default uses 10 uniform Euler steps from $\tau=1$ to $\tau=0$.

At every integration step, the 3D stream, semantic stream, and action expert are recomputed from the current action state.
Evolving action hypotheses inform future three-dimensional motion and semantic representations, allowing world predictions to continuously respond to candidate actions.
Updated 3D and semantic hidden states feed back to the action expert through layer-aligned interaction, informing action velocity-field predictions.
The integrator updates the action state accordingly and repeats the joint world--action computation at the next step.
Thus, the bidirectional interaction defined in Section~\ref{sec:world-action-comodeling} operates throughout continuous action generation, beyond its role as a training-time auxiliary constraint.

Because the VLM does not attend to the 3D, semantic, or action streams, its context remains unchanged within a flow integration process.
It is therefore computed and cached once at the start of each decision.
After integration, the generated action chunk is denormalized, converted from relative to absolute targets, and mapped to the target embodiment's control space for execution.
The policy then makes a new decision from updated observations.
Inference does not load the teacher models used for world supervision.
All world representations are generated directly by the internal 3D and semantic streams.

\section{Experiments}
\label{sec:experiments}

We conducted simulation evaluations, real-robot experiments, and mechanistic analyses to assess Magic-W0's task execution, downstream adaptation, and world--action interaction.
RoboDojo-Sim and LIBERO~\citep{liu2023libero} evaluated multitask manipulation, long-horizon tasks, and performance under scene randomization.
Five downstream real-robot manipulation tasks assessed adaptation of the pre-trained model through fine-tuning.
Mechanistic analyses further examined the dependence of world prediction on action inputs and the role of shared 3D representations in future semantic prediction.

\subsection{Simulation Experiments}
\label{sec:simulation-experiments}

\subsubsection{RoboDojo}
\label{sec:robodojo-evaluation}

\begin{figure}[!htbp]
    \centering
    \begin{subfigure}[t]{0.188\linewidth}
        \centering
        \includegraphics[width=\linewidth]{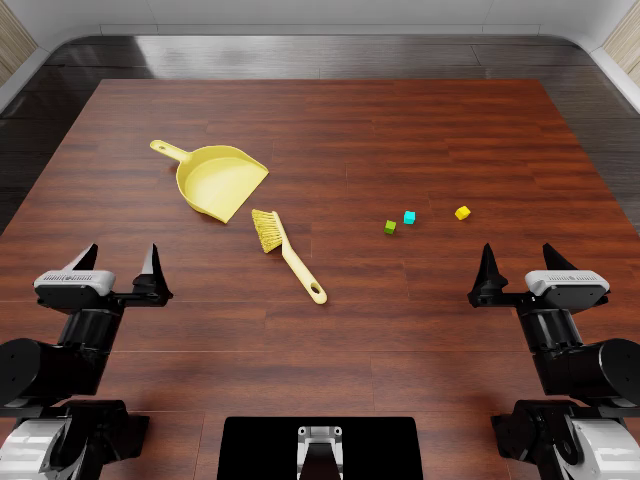}
        \caption{\textbf{Generalization}}
    \end{subfigure}\hfill
    \begin{subfigure}[t]{0.188\linewidth}
        \centering
        \includegraphics[width=\linewidth]{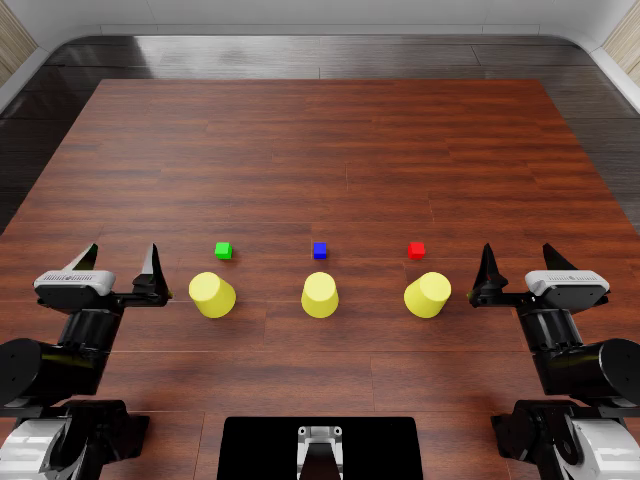}
        \caption{\textbf{Memory}}
    \end{subfigure}\hfill
    \begin{subfigure}[t]{0.188\linewidth}
        \centering
        \includegraphics[width=\linewidth]{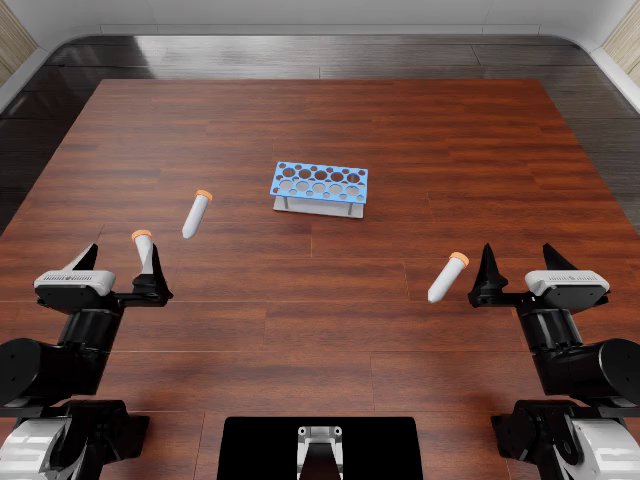}
        \caption{\textbf{Precision}}
    \end{subfigure}\hfill
    \begin{subfigure}[t]{0.188\linewidth}
        \centering
        \includegraphics[width=\linewidth]{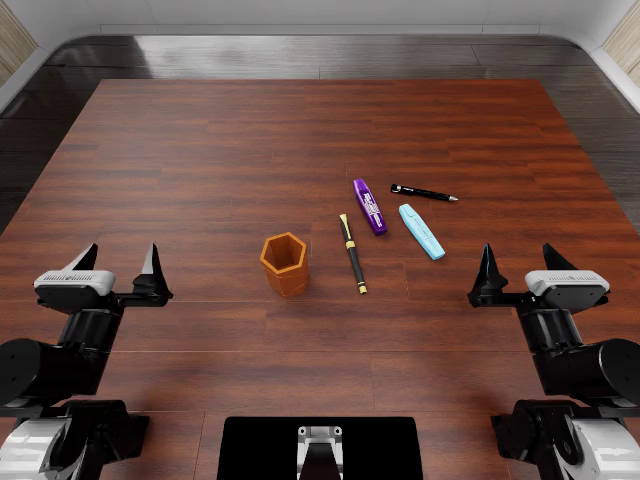}
        \caption{\textbf{Long-Horizon}}
    \end{subfigure}\hfill
    \begin{subfigure}[t]{0.188\linewidth}
        \centering
        \includegraphics[width=\linewidth]{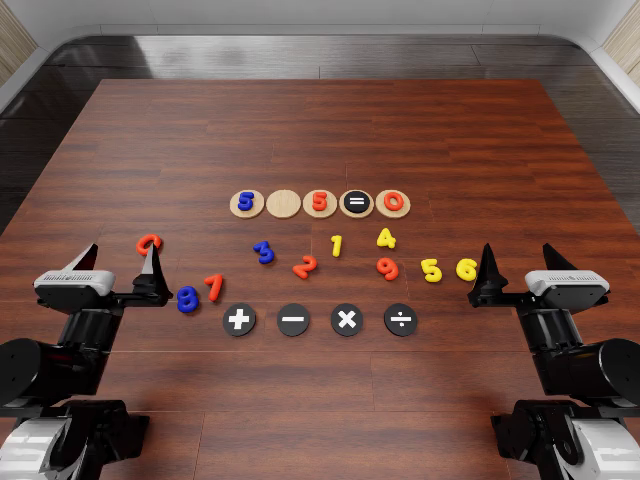}
        \caption{\textbf{Open}}
    \end{subfigure}
    \caption{\textbf{Representative RoboDojo-Sim tasks.} The five panels illustrate Generalization, Memory, Precision, Long-Horizon, and Open capabilities. Frames are taken from official RoboDojo task demonstrations~\citep{chen2026robodojo,robodojoTasks2026}.}
    \label{fig:robodojo-sim-tasks}
\end{figure}

RoboDojo-Sim~\citep{robodojoLeaderboard2026} comprises 42 simulation tasks and summarizes policy capabilities along five dimensions: Generalization, Precision, Long-Horizon, Memory, and Open.
Table~\ref{tab:robodojo-results} summarizes results for agents, VLAs, and WAMs.
Each dimension reports a process Score followed by success rate (SR, \%); Generalization averages the Gen-Std and Gen-Rand splits.
We evaluated Magic-W0 using the same tasks, observations, and execution protocol.
Its formal results were aggregated solely from complete official evaluation trajectories.
Figure~\ref{fig:robodojo-sim-tasks} shows representative tasks for the five evaluation dimensions.

\begin{table}[!htbp]
    \centering
    \caption{\textbf{RoboDojo-Sim comparison.} Each entry reports Score/SR (\%). Agent-based methods precede the VLA and WAM groups; VLA + WAM is included in the WAM group. VLA and WAM entries are sorted by ascending Average Score, with Magic-W0 listed last. Yes indicates that both code and model weights are publicly available. The highest Score in each column is bold. Leaderboard results were accessed on October 3, 2026.}
    \label{tab:robodojo-results}
    \tablefont
    \renewcommand{\arraystretch}{1.08}
    \setlength{\tabcolsep}{2.3pt}
    \begin{tabular}{llccccccc}
        \toprule
        Method & Type & \makecell{Open-\\source} & Generalization & Precision & \makecell{Long-\\Horizon} & Memory & Open & Average \\
        \midrule
        GPT-6-Astra~\citep{openai2026gpt6astra}            & Agent & No & 33.36/30.50 & 12.65/4.00  & 21.45/8.25  & 43.04/38.67 & \textbf{34.36}/31.00 & 28.97/22.48 \\
        PhysicalRSI~\citep{robodojoLeaderboard2026} & \makecell{Agent +\\VLA} & No & 21.30/15.62 & 38.05/32.50 & 46.37/37.33 & 46.74/46.56 & 28.88/24.92 & 36.27/31.38 \\
        \midrule
        VLAct~\citep{robodojoLeaderboard2026}                & VLA & Yes & 9.54/6.28  & 20.57/15.17 & 20.12/13.67 & 0.66/0.56   & 2.37/2.25  & 10.65/7.58 \\
        StarVLA-PI\_v3~\citep{robodojoLeaderboard2026}     & VLA & Yes & 11.22/8.05 & 17.77/12.50 & 18.46/11.00 & 4.59/4.00   & 2.03/2.00  & 10.81/7.51 \\
        InternVLA-A1.5~\citep{robodojoLeaderboard2026}     & VLA & Yes & 10.35/6.83 & 15.23/10.17 & 23.80/13.75 & 4.93/3.56   & 1.43/1.42  & 11.15/7.14 \\
        $\pi_{0.5}$~\citep{robodojoLeaderboard2026}         & VLA & Yes & 13.38/8.17  & 12.40/5.50  & 23.54/14.67 & 5.89/4.67   & 1.98/1.67  & 11.44/6.93 \\
        Spatial Forcing~\citep{robodojoLeaderboard2026}      & VLA & Yes & 14.12/9.34 & 17.32/10.58 & 23.26/14.58 & 5.43/4.11   & 1.78/1.58  & 12.38/8.04 \\
        SimpleMemVLA~\citep{robodojoLeaderboard2026}        & VLA & Yes & 6.36/3.95  & 7.42/2.92   & 14.58/5.50  & 33.71/33.22 & 0.85/0.75  & 12.58/9.27 \\
        KinRT~\citep{yang2026kinrt}                          & VLA & Yes & 14.02/8.61  & 15.65/9.92  & 26.40/18.08 & 4.82/3.56   & 4.23/3.83  & 13.02/8.80 \\
        Hy-Embodied-0.5-VLA~\citep{zhang2026hyvla}          & VLA & Yes & 11.78/8.39  & 13.81/8.00  & 25.74/14.92 & 13.37/12.11 & 0.65/0.58  & 13.07/8.80 \\
        Meituan-Robotics-0~\citep{robodojoLeaderboard2026} & VLA & Yes & 13.75/8.17  & 16.77/7.75  & 29.61/18.58 & 10.06/8.89  & 4.54/4.25  & 14.95/9.53 \\
        Xiaomi-Robotics-1~\citep{xiaomi2026robotics1}       & VLA & Yes & 23.54/17.00 & 26.69/18.83 & 38.39/23.67 & 7.81/6.56   & 3.94/3.58  & 20.07/13.93 \\
        GalaxeaVLA (G0.5)~\citep{galaxea2026g05}            & VLA & Yes & 18.46/12.83 & 28.25/20.42 & 44.12/32.25 & 8.61/7.33   & 1.73/1.58  & 20.23/14.88 \\
        DM0.5~\citep{dexmal2026dm05}                         & VLA & Yes & 15.77/10.95 & 24.82/16.75 & 33.70/19.50 & 47.74/47.44 & 2.43/2.08  & 24.90/19.34 \\
        Simate-beta~\citep{robodojoLeaderboard2026}      & VLA & No & 35.09/27.95 & 34.35/26.92 & \textbf{57.84}/43.42 & 33.33/33.00 & 9.12/8.50 & 33.95/27.96 \\
        \midrule[\heavyrulewidth]
        Fast-WAM~\citep{yuan2026fastwam}                     & WAM & Yes & 2.34/1.11  & 1.96/0.00   & 9.14/5.17   & 3.55/3.44   & 0.42/0.42  & 3.48/2.03 \\
        AHA-WAM~\citep{cai2026ahawam}                        & WAM & Yes & 5.79/3.28  & 5.86/2.42   & 8.61/2.67   & 2.97/2.78   & 0.88/0.83  & 4.82/2.39 \\
        GigaWorld-Policy-0~\citep{ye2026gigaworldpolicy}    & WAM & No & 5.35/2.89  & 6.15/1.83   & 15.51/8.92  & 3.46/2.22   & 0.54/0.50  & 6.20/3.27 \\
        X-WAM~\citep{guo2026xwam}                            & WAM & Yes & 7.39/3.33  & 6.72/1.83   & 17.47/9.08  & 6.32/4.67   & 0.57/0.25  & 7.69/3.83 \\
        OpenWAM-$\alpha$~\citep{wang2026openwam}             & WAM & Yes & 20.71/14.83 & 18.45/9.25  & 34.93/25.33 & 10.41/9.11  & 1.41/1.08  & 17.18/11.92 \\
        ME-U0~\citep{wen2026machembodiedu0}                 & WAM & No & 17.53/10.22 & 23.95/15.42 & 36.98/22.33 & 8.42/7.00   & 1.41/0.92  & 17.66/11.18 \\
        ME-Brain-1.0~\citep{robodojoLeaderboard2026} & \makecell{VLA +\\WAM} & Yes & 21.81/15.34 & 24.23/15.92 & 32.96/20.58 & 23.53/22.78 & 5.83/5.33 & 21.67/15.99 \\
        InternW0-$\Delta$~\citep{robodojoLeaderboard2026} & WAM & Yes & 30.09/22.78 & 31.98/23.25 & 46.29/29.33 & 34.67/34.00 & 10.84/10.17 & 30.77/23.91 \\
        VPP2-Preview~\citep{robodojoLeaderboard2026} & WAM & No & 31.34/24.62 & 33.75/25.58 & 35.31/22.17 & 50.78/50.33 & 5.83/5.42 & 31.40/25.62 \\
        Awomo-0.5~\citep{robodojoLeaderboard2026} & WAM & No & 30.31/23.84 & \textbf{39.65}/33.50 & 40.36/26.75 & 42.27/41.22 & 24.09/22.92 & 35.34/29.64 \\
        \midrule
        \textbf{Magic-W0} & WAM & Yes & \textbf{37.48}/29.17 & 37.14/30.73 & 51.80/36.00 & \textbf{52.68}/51.67 & 4.65/4.25 & \textbf{36.75}/30.36 \\
        \bottomrule
    \end{tabular}
\end{table}

Magic-W0 achieves the highest Average Score on RoboDojo-Sim, ranking No.~1 across all evaluated Agent+VLA, VLA+WAM, VLA, and WAM methods.
It reaches an Average Score of 36.75 with an average success rate of 30.36\%.
Compared with the strongest Agent+VLA method, PhysicalRSI, Magic-W0 improves the Average Score by 0.48 points; compared with the strongest WAM method, Awomo-0.5, by 1.41 points; and compared with the strongest VLA method, Simate-beta, by 2.80 points.
Magic-W0 also achieves the best overall performance in Generalization, with a Score of 37.48, and in Memory, with 52.68 Score / 51.67\% Success Rate, both ranking first among all compared methods.
Magic-W0's leading performance indicates that its representations extend beyond local visual patterns in current observations and use action-conditioned world-state changes for decision-making.
Current 3D Geometry, 3D Motion, and Future Semantics describe current physical structure, action-induced transitions, and future task states, respectively.
Layer-aligned world--action interaction incorporates this predictive information into action generation.
Structured world transition modeling therefore provides information that supports policy generalization under scene changes.

\subsubsection{LIBERO}
\label{sec:libero-evaluation}

To further assess downstream transfer, we fine-tuned Magic-W0 from its pre-trained weights on LIBERO and compared it with representative methods (Table~\ref{tab:libero-results}).
Baseline results are taken from published evaluations: Octo and OpenVLA from OpenVLA~\citep{kim2024openvla}; SpatialVLA from its original report~\citep{qu2025spatialvla}; $\pi_0$+FAST and OpenVLA-OFT from OpenVLA-OFT~\citep{kim2025openvlaoft}; GR00T-N1 and X-VLA from X-VLA~\citep{zheng2025xvla}; and $\pi_0$, $\pi_{0.5}$, Motus, LingBot-VA, and Fast-WAM from Fast-WAM~\citep{yuan2026fastwam}.
We retain the published values, including their reported averages.

Magic-W0 achieved an average SR of 99.1\%.
It obtained the highest SRs on Spatial and Goal tasks, at 99.4\% and 99.6\%, respectively.
Spatial performance is consistent with the joint modeling of three-dimensional geometry and vision--language context.
It indicates that explicit physical structure representations provide more effective spatial relationship information for the policy.
On Goal tasks, coupling 3D Motion and Future Semantics with action generation helps relate manipulation behaviors, environmental changes, and task goal states.
These results indicate that structured world representations transfer effectively to downstream manipulation, supporting reasoning and control across task types.

\begin{table}[!htbp]
    \centering
    \caption{\textbf{Comparison on LIBERO.}}
    \label{tab:libero-results}
    \tablefont
    \setlength{\tabcolsep}{7pt}
    \begin{tabular}{lccccc}
        \toprule
        Method & Spatial & Object & Goal & Long & Avg. SR \\
        \midrule
        Octo~\citep{ghosh2024octo} & 78.9 & 85.7 & 84.6 & 51.1 & 75.1 \\
        OpenVLA~\citep{kim2024openvla} & 84.7 & 88.4 & 79.2 & 53.7 & 76.5 \\
        SpatialVLA~\citep{qu2025spatialvla} & 88.2 & 89.9 & 78.6 & 55.5 & 78.1 \\
        GR00T-N1~\citep{nvidia2025gr00tn1} & 94.4 & 97.6 & 93.0 & 90.6 & 93.9 \\
        $\pi_0$+FAST~\citep{pertsch2025fast} & 96.4 & 96.8 & 88.6 & 60.2 & 85.5 \\
        $\pi_0$~\citep{black2024pi0} & 96.8 & 98.8 & 95.8 & 85.2 & 94.1 \\
        $\pi_{0.5}$~\citep{black2025pi05} & 98.8 & 98.2 & 98.0 & 92.4 & 96.9 \\
        OpenVLA-OFT~\citep{kim2025openvlaoft} & 97.6 & 98.4 & 97.9 & 94.5 & 97.1 \\
        X-VLA~\citep{zheng2025xvla} & 98.2 & 98.6 & 97.8 & 97.6 & 98.1 \\
        Motus~\citep{bi2025motus} & 96.8 & 99.8 & 96.6 & 97.6 & 97.7 \\
        LingBot-VA~\citep{li2026causalworld} & 98.5 & 99.6 & 97.2 & \textbf{98.5} & 98.5 \\
        Fast-WAM~\citep{yuan2026fastwam} & 98.2 & \textbf{100.0} & 97.0 & 95.2 & 97.6 \\
        \midrule
        \textbf{Magic-W0} & \textbf{99.4} & 99.2 & \textbf{99.6} & 98.0 & \textbf{99.1} \\
        \bottomrule
    \end{tabular}
\end{table}

\FloatBarrier
\subsection{Real-Robot Experiments}
\label{sec:downstream-finetuning}

\begin{figure}[!htbp]
    \centering
    \includegraphics[width=\linewidth]{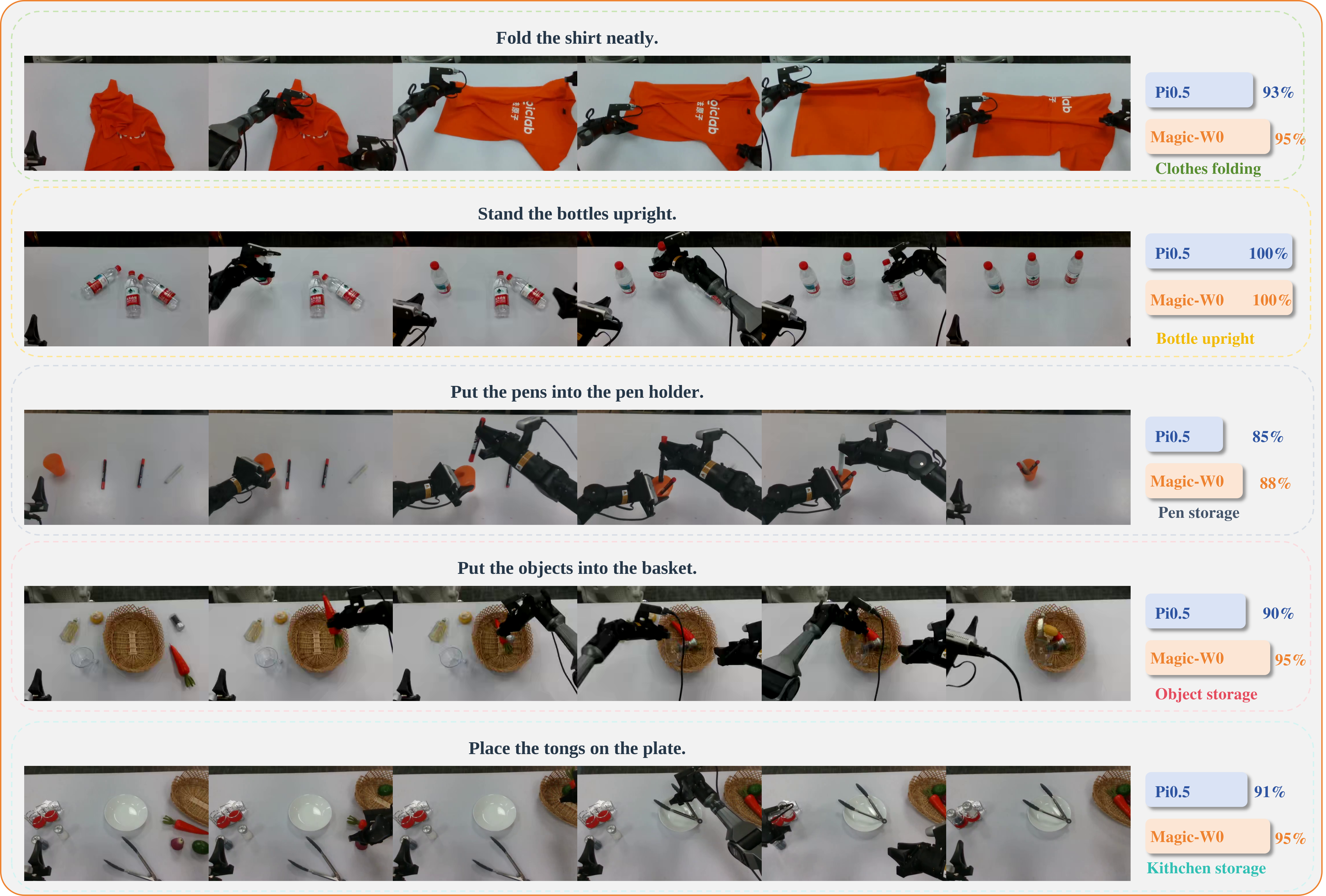}
    \caption{\textbf{Video keyframes and success rates for five real-robot tasks.} Rows show six keyframes each for clothes folding, bottle uprighting, pen storage, object storage, and kitchen storage. Task SRs for $\pi_{0.5}$ and Magic-W0 appear on the right. Row titles are English instructions composed from the task content and are not the original prompts used in the experiments.}
    \label{fig:real-robot-sequences}
\end{figure}

To assess downstream adaptation and execution on real robots, we fine-tuned and evaluated Magic-W0 on five new manipulation tasks.
These were clothes folding, bottle uprighting, pen storage, object storage, and kitchen storage.
They cover deformable-object manipulation, precise pose control, long-horizon organization, and language-conditioned manipulation in scenes with multiple objects.
For fair comparison, Magic-W0 and $\pi_{0.5}$ used identical demonstrations, training budgets, visual and proprioceptive observations, action spaces, and initial evaluation states.
Each task was evaluated over 100 independent trials, with task SR as the primary metric.
Figure~\ref{fig:real-robot-sequences} shows execution keyframes and corresponding SRs.
Magic-W0 achieved an average SR of 94.6\%, exceeding $\pi_{0.5}$'s 91.8\% by 2.8 percentage points.
It improved on four tasks and matched $\pi_{0.5}$'s 100\% SR on bottle uprighting, with no task-level performance decrease.
The largest improvements occurred in object storage, from 90\% to 95\%, and kitchen storage, from 91\% to 95\%.
Pen storage and clothes folding improved by three and two percentage points, respectively.
Both methods reached saturated performance on bottle uprighting, leaving no further difference on that task.

Magic-W0's advantage was more evident in manipulation requiring sustained tracking of environmental changes.
Object storage and kitchen storage involve multiple objects and successive state transitions, requiring policies to adjust later actions to scene changes caused by earlier operations.
Clothes folding additionally involves continuous changes in deformable-object shape.
Consistent improvements on these tasks align with Magic-W0's structured world modeling design.
The model jointly represents current physical states, action-induced changes, and future states through Current 3D Geometry, 3D Motion, and Future Semantics.
This provides predictive world information for continuous action generation alongside current observations.

\subsection{World--Action Interaction Analysis}
\label{sec:world-action-analysis}

Layer-aligned cross-stream interaction provides explicit pathways for action information to reach the structured world heads.
We designed two intervention experiments to test whether these pathways are used during inference.
First, does changing action conditions systematically affect structured world outputs when observations, tasks, and proprioceptive states remain fixed?
Second, how do different connections among the action, 3D, and semantic streams affect Future Semantics?
We examined these questions through cross-sample action replacement and cross-stream connection masking, respectively.

All experiments used six batches, each containing four observation times sampled from trajectories.
Within each comparison, we fixed current observations, language instructions, proprioceptive states, and flow time $\tau$, and reused the same noise samples.
Following Equation~\eqref{eq:flow-path}, the noisy action chunk is
\begin{equation}
    \mathbf x_\tau=(1-\tau)\action+\tau\bm{\epsilon},
\end{equation}
where $\action$ is the ground-truth action chunk and $\bm{\epsilon}$ is Gaussian noise.
As $\tau$ increases, $\mathbf x_\tau$ retains less ground-truth action information.
We divided $\tau$ into five equal intervals and summarized results separately within each interval.

We measured each structured world head's response by the feature-space distance between its outputs before and after intervention.
For Future Semantics, $\Loss_{\mathrm{sem}}$ additionally measured the difference from teacher features extracted from the original sample's future observation.
Lower loss indicates closer agreement with the ground-truth future representation.
Unless otherwise stated, error bars denote standard deviations across the six batches.

\Needspace{6\baselineskip}
\subsubsection{Action Sensitivity}
\label{sec:action-sensitivity}

We first examined whether the structured world heads use action conditions.
For sample $i$, we kept current observations, language instructions, proprioceptive states, and flow time $\tau$ fixed.
We replaced only its paired noisy action chunk $\mathbf x_\tau^{(i)}$ with that of another sample in the same batch:
\begin{equation}
    \tilde{\mathbf x}^{(i)}_\tau=\mathbf x^{(\pi(i))}_\tau,
    \qquad \pi(i)\neq i ,
\end{equation}
where $\pi$ denotes reassignment of samples within the batch.
This intervention changes the action condition while preserving the scene, task, and proprioceptive state.
We refer to it as \emph{action replacement}.
If the structured world heads depend on action conditions, replacing actions alone should measurably change Current 3D Geometry, 3D Motion, or Future Semantics.

The three heads operate in different feature spaces and at different output scales.
We therefore normalized intervention magnitudes by the mean feature distance between different samples within the batch:
\begin{equation}
D_m(\tau)=
\frac{
\mathbb{E}_i
\left\|
\mathbf Z_m^{(i)}(\mathbf x_\tau^{(i)})
-
\mathbf Z_m^{(i)}(\tilde{\mathbf x}_\tau^{(i)})
\right\|
}{
\mathbb{E}_{i\neq j}
\left\|
\mathbf Z_m^{(i)}(\mathbf x_\tau^{(i)})
-
\mathbf Z_m^{(j)}(\mathbf x_\tau^{(j)})
\right\|
},
\qquad
m\in\{\mathrm{geo},\mathrm{mot},\mathrm{sem}\}.
\label{eq:action-sensitivity}
\end{equation}
$D_m=0$ indicates that action replacement barely changes the head's output.
$D_m=1$ indicates a change as large as the mean feature distance between different samples within the batch.

We also compared $\Loss_{\mathrm{sem}}$ before and after action replacement.
Beyond measuring output changes, this metric tests whether matching actions to the original future state affects Future Semantics prediction quality.
As a reference, we also report the normalized change in the action expert's velocity-field output under the same intervention.

\begin{figure}[!htbp]
\centering
\includegraphics[width=\linewidth]{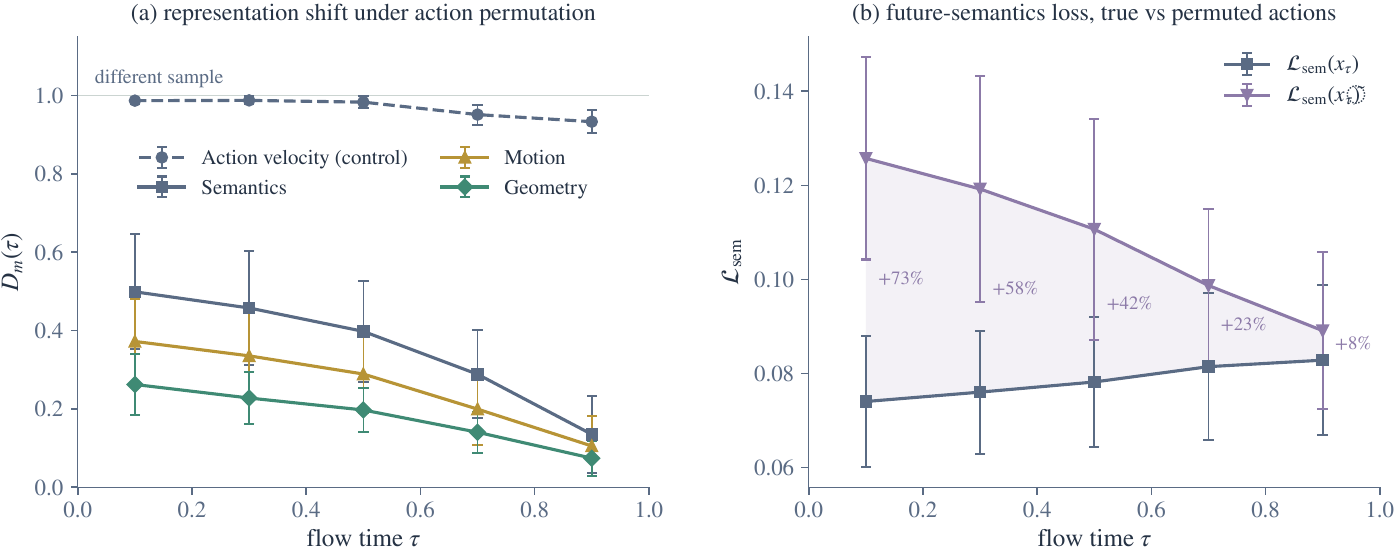}
\caption{\textbf{Effects of action replacement on structured world heads.}
(a) Normalized output changes $D_m(\tau)$ for the three heads. The horizontal reference line represents the mean output distance between samples within a batch; the dashed line shows the action expert velocity-field control.
(b) Future Semantics loss before and after action replacement. Percentages are batch-averaged values of $\Loss_{\mathrm{sem}}(\tilde{\mathbf x}_\tau)/\Loss_{\mathrm{sem}}(\mathbf x_\tau)-1$. Larger $\tau$ indicates less ground-truth action information in $\mathbf x_\tau$. Error bars denote standard deviations across batches.}
\label{fig:world-action-sensitivity}
\end{figure}

As shown in Figure~\ref{fig:world-action-sensitivity}, changing action conditions alone altered all three structured world outputs, with responses decreasing as $\tau$ increased.
At $\tau\in[0,0.2)$, $D_m$ was 0.50 for Future Semantics, 0.37 for 3D Motion, and 0.26 for Current 3D Geometry.
At $\tau\in[0.8,1.0)$, these values decreased to 0.14, 0.11, and 0.07, respectively.
The corresponding differences between the intervals were 0.36, 0.26, and 0.19.
Since $\mathbf x_\tau$ retains less ground-truth action information at larger $\tau$, this trend links the heads' responses to the amount of available action information.
By comparison, the action expert's velocity-field output responded strongly across all noise intervals, with normalized changes remaining between 0.93 and 0.99.

Future Semantics prediction loss provides additional evidence.
In the lowest-noise interval, action replacement increased $\Loss_{\mathrm{sem}}$ by 73.5\%.
This relative increase declined with $\tau$, reaching 8.0\% in the highest-noise interval, where the mean $\pm$ one across-batch standard deviation included zero.
Without action replacement, $\Loss_{\mathrm{sem}}$ itself increased from 0.0741 to 0.0828 as action noise increased.

These observations show that both output representations and prediction quality depend on action conditions.
With visual observations, language instructions, and proprioceptive states fixed, changing action inputs alone systematically altered the structured world heads' outputs.
For Future Semantics, this mismatch also increased prediction loss.
Both effects weakened as ground-truth action information in $\mathbf x_\tau$ decreased.
The results indicate that Structured World Transition uses action conditions to form world representations, rather than deriving its outputs solely from VLM context.

\Needspace{6\baselineskip}
\subsubsection{Cross-Stream Dependence}
\label{sec:cross-stream-dependence}

Action replacement shows that structured world outputs respond to action conditions, but it does not distinguish the internal pathways influencing Future Semantics.
We therefore directly intervened in information transfer between expert streams in the second experiment.
Inputs and all other computations remained unchanged; only the specified directed cross-stream connection was masked.

Let $u\rightarrow m$ denote target stream $m$ attending to intermediate representations from source stream $u$ through joint attention.
This corresponds to row $m$, column $u$ in the visibility matrix of Figure~\ref{fig:architecture}.
For example, masking $3d\rightarrow sem$ prevents the semantic stream from attending to 3D keys and values.
Within-stream computation and access to other visible sources remain intact.
As an overall control, the \emph{parallel} configuration removes all connections between expert streams simultaneously.
Each expert then uses only shared VLM context and its own within-stream representations.

To assess the contribution of each pathway to Future Semantics, we measured changes in $\Loss_{\mathrm{sem}}$.
The full model's mean semantic loss was 0.0785.
For each masking configuration, we calculated the increase in $\Loss_{\mathrm{sem}}$ relative to the full model.

\begin{figure}[!htbp]
\centering
\includegraphics[width=\linewidth]{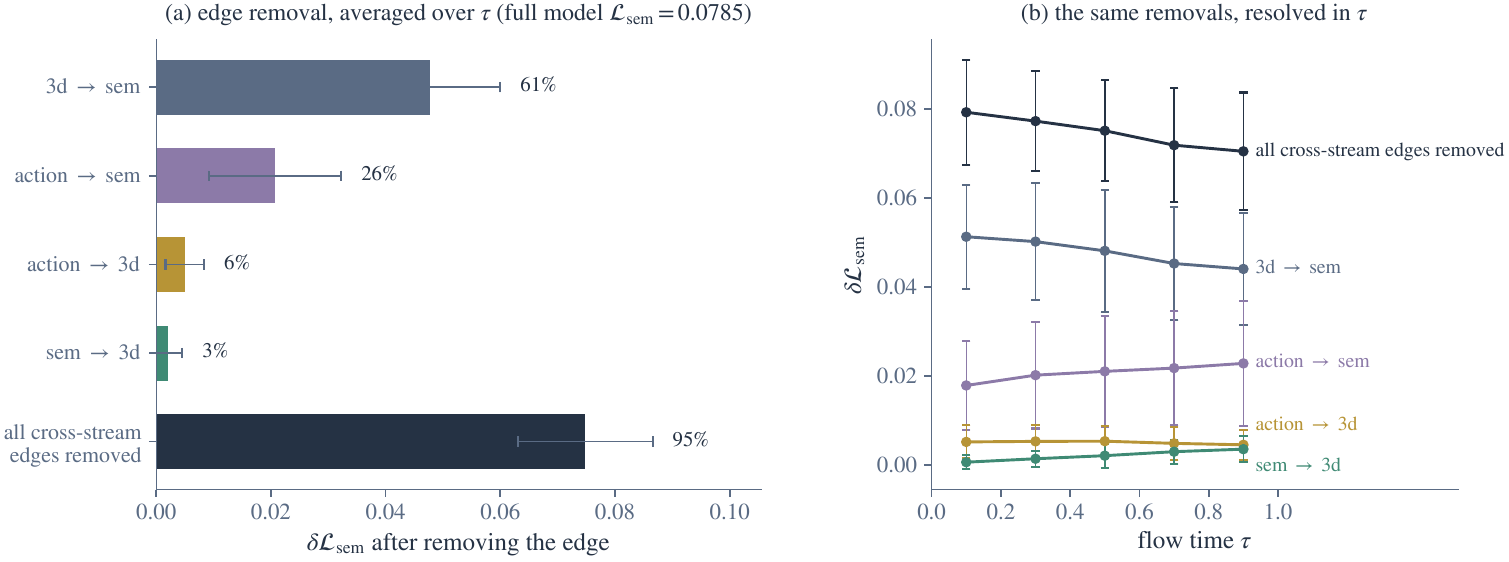}
\caption{\textbf{Effects of cross-stream connection masking on Future Semantics.}
(a) Semantic loss increases after masking different connections, averaged over all flow intervals. Percentages are relative to the full model.
(b) Results for the same interventions at different flow times $\tau$. Arrows indicate information flow; \emph{all cross-stream edges removed} denotes simultaneous removal of all connections between expert streams. Error bars denote standard deviations across batches.}
\label{fig:world-stream-edges}
\end{figure}

Figure~\ref{fig:world-stream-edges} shows pronounced asymmetry in how directed connections affect Future Semantics.
Masking $3d\rightarrow sem$ increased semantic loss by 60.8\%, the largest effect among the individual connections examined.
Masking the reverse connection, $sem\rightarrow3d$, increased loss by only 2.6\%.
Under this metric, the former increase was approximately 23 times the latter.
This indicates strong dependence of semantic prediction on 3D information, with a weaker direct effect of reverse information transfer on this metric.
Since Current 3D Geometry and 3D Motion share 3D hidden states, the $3d\rightarrow sem$ connection provides representations constrained by both geometry and motion supervision.

Action information influences Future Semantics through direct and indirect pathways.
Masking $action\rightarrow sem$ directly removes the semantic stream's access to action representations.
Masking $action\rightarrow3d$ blocks the route through which action information first enters the 3D stream and then reaches the semantic stream via $3d\rightarrow sem$.
Both interventions increased $\Loss_{\mathrm{sem}}$.
Future Semantics therefore uses action information directly and also depends on action-related information conveyed through the 3D stream.

Removing all cross-stream connections increased semantic loss by 95.3\% relative to the full model.
The loss increases from individually masking the four examined connections summed to 0.0755.
Removing all cross-stream connections simultaneously increased loss by 0.0748, a difference of approximately 1.0\%.
This suggests that the full model's Future Semantics advantage over the parallel configuration is mainly associated with these information pathways.
However, different connections may affect shared intermediate representations.
This numerical relationship should therefore not be interpreted as strict additivity of individual connection contributions.

Across $\tau$ intervals, the loss increase from masking $3d\rightarrow sem$ varied relatively little with noise level, while the effect of masking $action\rightarrow sem$ increased slightly.
The two experiments intervene on different components.
Section~\ref{sec:action-sensitivity} changes action inputs while preserving all internal information pathways; this section fixes inputs and removes specific pathways.
Their trends across $\tau$ therefore need not coincide.

Together, the experiments provide two complementary observations.
First, structured world outputs are not determined solely by VLM context; they change systematically with action conditions.
Second, Future Semantics clearly depends on cross-stream information from both the 3D and action streams, particularly the $3d\rightarrow sem$ connection.
These findings indicate that layer-aligned cross-stream interaction participates in computing action-conditioned Structured World Transition.

\Needspace{9\baselineskip}
\section{Conclusion}
\label{sec:conclusion}

We presented \textbf{Magic-W0}, a structured world--action foundation model for robot manipulation.
Magic-W0 jointly learns structured world evolution and robot action generation from large-scale heterogeneous data across sources, applying world prediction to robot manipulation.
Unlike VLAs that generate actions solely from observations, Magic-W0 expresses environmental evolution during robot interaction as \textbf{Current State--Transition--Future State}.
\textbf{Structured World Transition} represents the current state, action-conditioned three-dimensional state transitions, and task-relevant future states.
To unify world prediction and action generation, Magic-W0 uses a layer-aligned world--action interaction architecture.
Structured world and action representations interact continuously at multiple network depths, bidirectionally coupling \textbf{action-conditioned world transition} and \textbf{world-informed action generation}.
The model learns cross-embodiment transferable robot representations through a unified state--action interface.
Training uses heterogeneous data spanning egocentric human manipulation, UMI, real robots, and simulation, with latent supervision from frozen geometry, motion, and semantic vision models.
On RoboDojo-Sim, Magic-W0 achieved an average Score of 36.75, ranking first among all compared methods.
On multiple real-robot tasks, the model performed well after fine-tuning on limited downstream data.
Further world--action interaction diagnostics showed that its learned structured world representations adjusted to changes in candidate actions.
They also showed that shared three-dimensional representations conveyed action-related information to task-relevant future state predictions.
Overall, Magic-W0 explores a robot foundation model paradigm that moves from action imitation toward joint modeling of world understanding and action generation.
It offers a new path toward general-purpose embodied intelligence with physical world perception, prediction, and interaction capabilities.

Although Magic-W0 demonstrates the potential of structured world modeling for robot manipulation, several directions remain to be explored.
We plan to extend the framework to \textbf{dexterous-hand precision manipulation}.
This will examine its modeling and generalization in high-dimensional action spaces, fine-grained contact interactions, and complex manipulation scenarios.
We also plan to incorporate \textbf{tactile and force signals} to more fully characterize robot--environment contact states and interaction dynamics.
Finally, we aim to address the additional computational cost of multilayer world--action interaction and \textbf{narrow the inference-speed gap with state-of-the-art VLA models}.

\section*{Contributions and Acknowledgments}
\phantomsection
\addcontentsline{toc}{section}{Contributors}
\begingroup
\small
\setstretch{1.18}
\setlength{\parindent}{0pt}
The following people contributed to Magic-W0.

\vspace{0.6em}
\noindent\textbf{Core Contributors:}

Xuhua Chen, Zhenhan Yin, Yuan Zhang, and Tao Zhang\textsuperscript{\scalebox{1.3}{\ensuremath{\bm{\ast}}}}.

\vspace{0.6em}
\noindent\textbf{Contributors:}

Lingfeng Zhang, He Zheng, Tong Mu, Shun Zuo, Dian Zhou, Di Wu, Xuan Zhou, Shaojie Wan, Rongtian Shen, Qiulong Xu, Yiduo Li, Yinglong Wang, Yanqian Wang, and Kun Wang.

\vspace{0.6em}
\noindent\textbf{Acknowledgements:}

We thank Zhenghua Xie, Shiyi Zhu, Yan Zhang, Xiaohui Wang, Lihao Han, Jian Zhou, Yadong Liu, Zehua Jiang, Lei Gao, Xiaonan He, Yuan Wang, Pengpeng Xu, Yue Zhao, and Hongliang Li for their support and contributions to this project. These contributions included data preparation, model evaluation, infrastructure support, and helpful discussions.

We also thank Professors Jiangning Zhang (Zhejiang University), Wenqi Zhang (Zhejiang University), and Xuanhan Wang (Tongji University) for their valuable advice and discussions.

\vspace{0.6em}
\noindent\textsuperscript{\scalebox{1.3}{\ensuremath{\bm{\ast}}}}\,Project Lead.
\endgroup

\bibliographystyle{plainnat}
\bibliography{references}

\clearpage
\appendix
\phantomsection
\addcontentsline{toc}{section}{Appendix}
\addtocontents{toc}{\protect\cftsetindents{subsection}{1em}{1.2em}%
    \protect\cftsetindents{subsubsection}{2.2em}{1.9em}}
\centerline{\Large\sffamily\bfseries Appendix}
\vspace{0.3em}
\begingroup
\let\magicmainaddcontentsline\addcontentsline
\renewcommand{\addcontentsline}[3]{%
    \ifstrequal{#1}{toc}{%
        \ifstrequal{#2}{section}{%
            \magicmainaddcontentsline{#1}{subsection}{#3}%
        }{%
            \ifstrequal{#2}{subsection}{%
                \magicmainaddcontentsline{#1}{subsubsection}{#3}%
            }{\magicmainaddcontentsline{#1}{#2}{#3}}%
        }%
    }{\magicmainaddcontentsline{#1}{#2}{#3}}%
}
\small
\setstretch{1.0}
\setlength{\abovedisplayskip}{6pt}
\setlength{\belowdisplayskip}{6pt}
\setlength{\abovedisplayshortskip}{4pt}
\setlength{\belowdisplayshortskip}{4pt}
\section{Pre-Training Data and Processing}
\label{app:data}

This section summarizes the pre-training corpora and implementation details of the unified state--action interface in Section~\ref{sec:cross-embodiment}.
These details cover coordinate transformations, supervision construction, and training sample construction.
Offline processing unifies physical units, left--right arm ordering, gripper definitions, and media indices, and associates camera calibration parameters with each trajectory.
Each processed trajectory contains task text, timestamps, multi-camera videos, states and actions, camera intrinsics and poses, and per-frame quality flags.

\subsection{Pre-Training Corpora}
\label{app:pretraining-corpora}

Tables~\ref{tab:embodied-pretraining-corpus} and~\ref{tab:vlm-sft-corpus} list the manipulation and vision--language supervised fine-tuning corpora used for pre-training, respectively.
Counts are approximate numbers of valid samples after filtering and aggregation; they should not be equated with the official total sizes of the datasets.
EO-Data1.5M and Robo2VLM-1 are independent open-source datasets, while in-house annotations are constructed from selected open-source datasets.

\begin{table}[!htbp]
\centering
\captionsetup{width=0.92\linewidth}
\caption{\textbf{Composition of the embodied pre-training corpus.}}
\label{tab:embodied-pretraining-corpus}
\tablefont
\begin{tabularx}{0.92\linewidth}{@{}lXr@{}}
\toprule
Data category & Dataset & Valid episodes \\
\midrule
\multirow{2}{*}{Egocentric human} & EgoSuite & 360k \\
 & EgoDex & 330k \\
\midrule
UMI & Hy-Embodied-0.5-VLA-Data & 231k \\
\midrule
\multirow{2}{*}{Simulation} & InternData-A1 & 374k \\
 & RoboTwin 2.0 & 73.2k \\
\midrule
\multirow{4}{*}{Real robot} & AgiBotWorld-Beta & 500k \\
 & RoboCOIN & 120k \\
 & AgiBot World 2026 & 20.7k \\
 & Galaxea Open-World Dataset & 5.2k \\
\bottomrule
\end{tabularx}
\end{table}

\begin{table}[!htbp]
\centering
\captionsetup{width=0.92\linewidth}
\caption{\textbf{Composition of the vision--language supervised fine-tuning corpus.}}
\label{tab:vlm-sft-corpus}
\tablefont
\begin{tabularx}{0.92\linewidth}{@{}lXr@{}}
\toprule
Data category & Dataset & Samples \\
\midrule
\multirow{2}{*}{Open-source datasets} & EO-Data1.5M & 1.42M \\
 & Robo2VLM-1 & 685k \\
\midrule
In-house annotations & Annotations constructed from open-source datasets & 500k \\
\bottomrule
\end{tabularx}
\end{table}

\FloatBarrier

\subsection{Cross-Source Supervision Construction}
\label{sec:retargeting}

\paragraph{Coordinate transformations and validity.}
Let \({}^{W}\!\mathbf T_{C_t}\) denote the camera pose in world coordinates at frame $t$, and \({}^{W}\!\mathbf T_E\) the end-effector pose at the same time.
Each frame's end-effector pose is transformed into that frame's camera coordinates:
\begin{equation}
{}^{C_t}\!\mathbf T_E=\left({}^{W}\!\mathbf T_{C_t}\right)^{-1}{}^{W}\!\mathbf T_E.
\label{eq:camera-frame-eef}
\end{equation}
Both poses must use the same reference frame and correspond to the same video frame.
Offline processing transforms each frame separately, rather than reusing one camera pose throughout an action chunk.
Rotations use a 6D representation, avoiding quaternion sign ambiguity and Euler-angle discontinuities.
Each arm has six rotation dimensions, with validity assessed separately for the two sides.
Every source records its world orientation, base frame, quaternion ordering, and end-effector axis definitions.
Figure~\ref{fig:eef-alignment} shows projection examples.

\begin{figure}[!htbp]
\centering
\includegraphics[width=\linewidth]{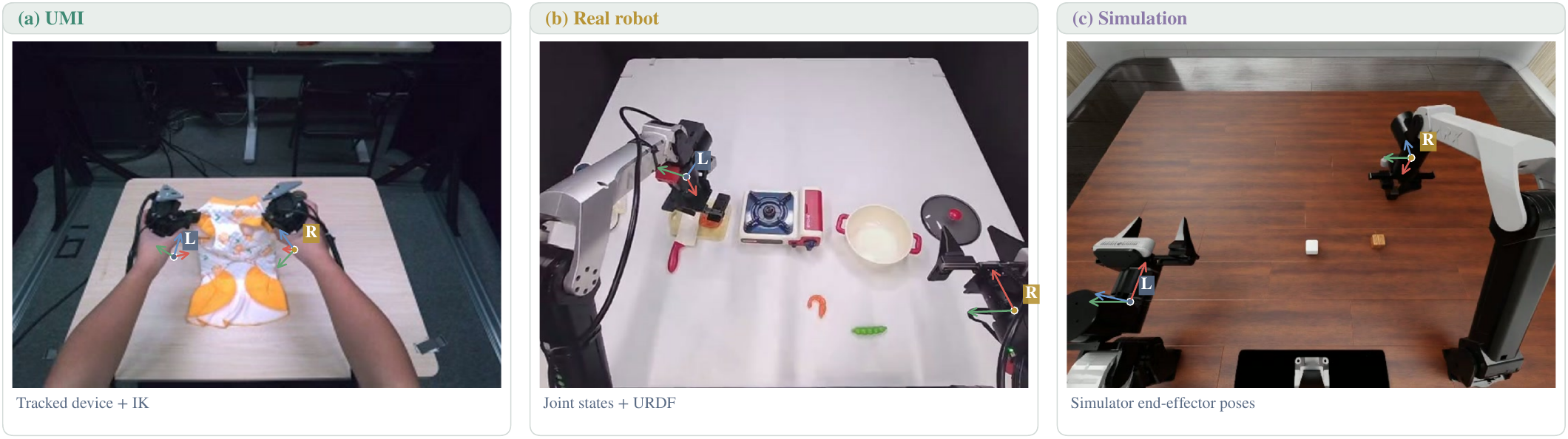}
\caption{\textbf{End-effector pose projections from three source categories under a unified convention.} Each source uses its own camera poses and intrinsics to project end-effector poses onto observations. UMI poses are obtained from tracked devices and inverse kinematics; real-robot poses from joint states and URDF; simulation poses directly from simulator records. Red, green, and blue indicate local $x$, $y$, and $z$ axes, respectively. Figure~\ref{fig:data-pipeline} shows projections for egocentric human manipulation.}
\label{fig:eef-alignment}
\end{figure}

\paragraph{Egocentric human manipulation supervision.}
Raw records in this category contain no robot proprioceptive states.
We construct virtual end-effector poses from hand keypoints and represent gripper opening using the distance between the thumb and middle finger.
Local axes are first unified across hands, then transformed into camera coordinates.
During pre-training, these trajectories are learned at the end-effector pose level without mapping them to a specific robot.
Downstream fine-tuning adapts the joint representation to the target robot.

\paragraph{UMI kinematic retargeting.}
Tracked device poses are registered to the robot base frame.
A rigid device-to-gripper transformation then yields target gripper poses.
Inverse kinematics (IK) solves for joint angles with joint-range and per-step change limits.
The solution provides joint and gripper supervision.
Each trajectory stores solver errors and collision-check results for subsequent filtering.
End-effector poses are retained in both original and camera coordinate frames.
Figure~\ref{fig:data-pipeline} shows the two construction pipelines.

\begin{figure}[!htbp]
\centering
\includegraphics[width=\linewidth]{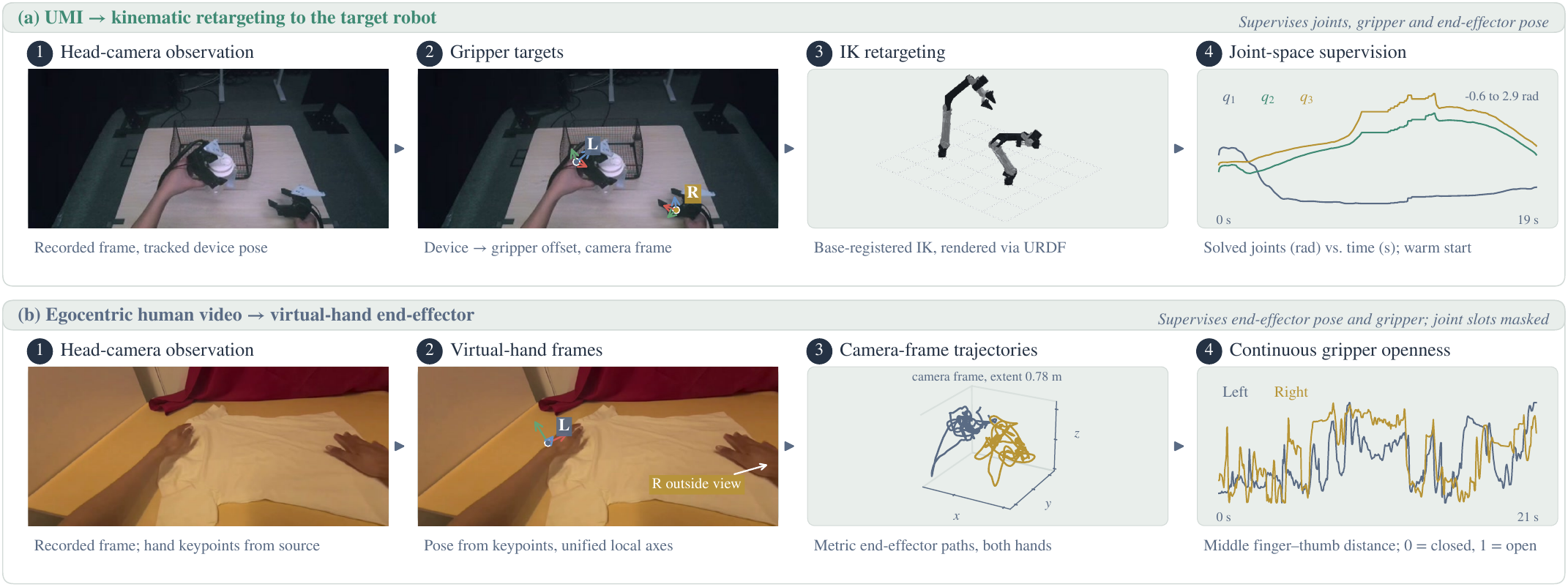}
\caption{\textbf{Raw UMI and egocentric human manipulation records contain no proprioceptive states for the target robot, requiring constructed supervision.} UMI tracked poses yield bimanual joint angles and gripper openings through IK, while retaining end-effector poses. Human manipulation videos use hand keypoints to derive virtual end-effector poses, camera-frame trajectories, and continuous gripper openings.}
\label{fig:data-pipeline}
\label{fig:ego-to-robot}
\end{figure}

\FloatBarrier
\subsection{Training Samples}
\label{app:training-samples}

Each training sample is constructed around anchor time $t$ as
\begin{equation}
\mathcal{B}_t=\left(
\obs_{\leq t},\lang,\state_t,
\action_{t:t+H-1},\obs_{t+\Delta},
\mathbf M_t^{\mathrm{state}},\mathbf M_{t:t+H-1}^{\mathrm{act}}\right),
\label{eq:data-sample}
\end{equation}
where $\mathbf M_t^{\mathrm{state}}$ and $\mathbf M_{t:t+H-1}^{\mathrm{act}}$ are state and action validity masks, respectively.
Anchor gating and future-timestep masks are configured separately for each source (Appendix~\ref{app:data-checks}).
An anchor can contribute supervision at only some timesteps.
World targets are not stored in advance with samples; they are extracted from temporally aligned observations during training forward passes.

Observations use three cameras: one head camera and two wrist cameras.
Anchors are sampled at a fixed stride of 25 frames, with additional anchors from keyframe annotations where available.
The stride is measured in frames; its duration depends on each source's native frame rate.

Offline processing stores absolute states and absolute control targets.
Native control targets are preserved; for trajectories without them, next-timestep states provide supervision.
The training loader constructs action chunks around each anchor.
Temporal resampling is performed in the absolute representation.
Joint angles and end-effector poses are then converted to quantities relative to the same anchor state, while grippers retain absolute openings.
Actions are scaled by source and dimension using the 1st and 99th percentiles, then clipped to $[-1,1]$.
All timesteps within a chunk share normalization statistics.
State and action validity masks are constructed together with each sample.

Future-frame offsets follow Equation~\eqref{eq:source-aware-horizon}, with action sampling multiplier $\rho_i$ configured by source.
Multipliers are typically greater than one for egocentric human manipulation and between $0.9$ and $1.2$ for teleoperated data.
For EgoDex, $\rho_i=1.95$ and $H=50$ yield $\Delta_i=26$.
Pre-training excludes anchors whose future targets extend beyond the episode endpoint.
Fine-tuning in Section~\ref{sec:finetuning} holds endpoint values: repeated terminal actions count as valid supervision, and world targets reuse the final frame.

\section{Data Processing and Quality Control}
\label{app:data-checks}
\label{sec:data-processing}

This section describes calibration when camera parameters are missing, source-specific quality checks, and how their outputs enter training.

\subsection{Camera Calibration for Sources without Metadata}
\label{sec:camera-calibration}

Camera-frame end-effector targets require camera intrinsics and poses for each record, but some sources do not release these parameters.
For these records, we estimate calibration from videos and available metric poses (Figure~\ref{fig:camera-calibration}).
For bimanual real-robot data, we use silhouette registration.
Temporal-median background subtraction extracts moving foreground regions, while joint states and URDF generate projected geometry for both arms.
We jointly estimate camera pose, focal length, link-width scale, and the transform between arm bases by maximizing overlap between projected silhouettes and foreground masks.
This method requires no calibration board, but depends on foreground extraction quality.
Calibration is therefore validated by spot-checking two-arm projections across frames and tasks.
For UMI, we fit correspondences between device trajectories and images.
Motion information within the tabletop region and low-intensity connected components identify device candidates.
These candidates are paired with tracked poses to establish three-dimensional-to-two-dimensional correspondences.
We jointly estimate camera pose, focal length, principal point, and device offset.
The offset is used only in calibration and does not alter the definition of training end-effector targets.
Calibration parameters are shared within validated recording groups; changes in head-mounted camera pose require regrouping and revalidation.

\begin{figure}[tbp]
\centering
\includegraphics[width=\linewidth]{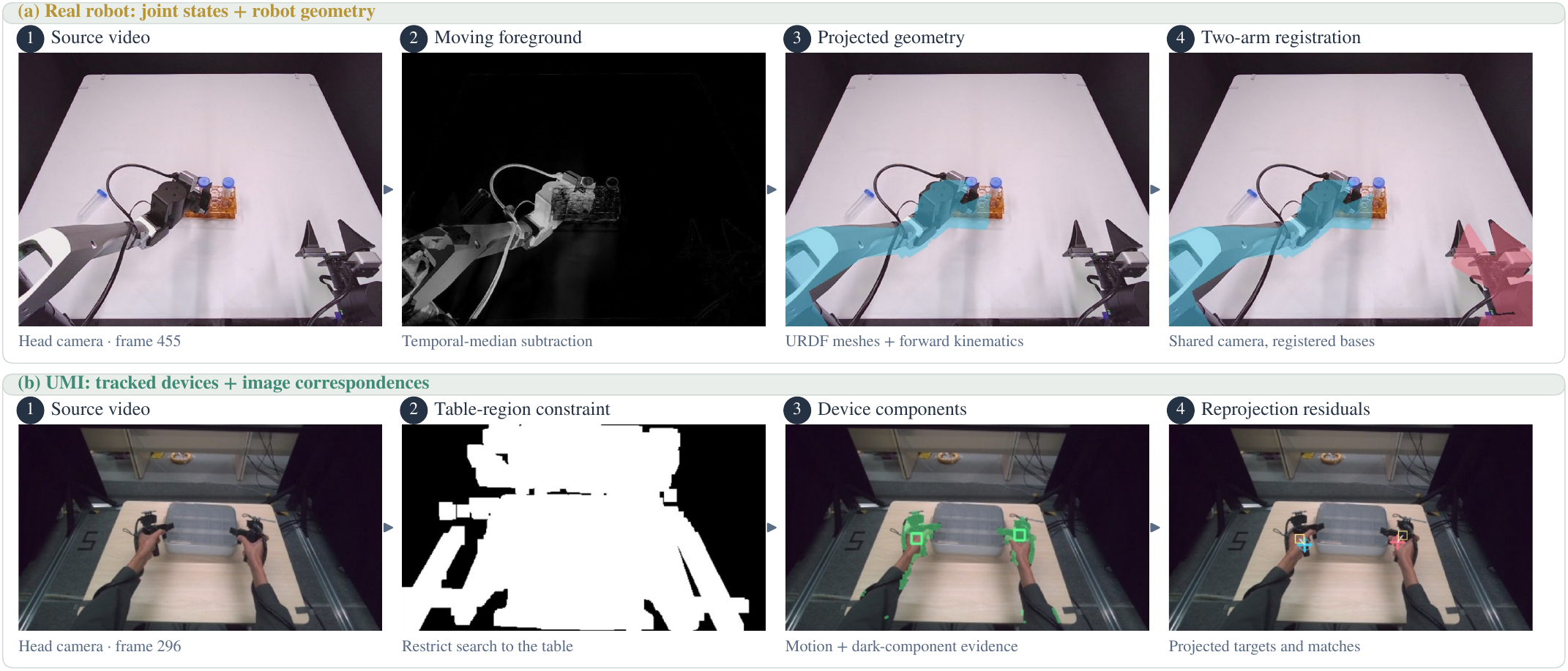}
\caption{\textbf{Calibration when camera parameters are unavailable.} Real-robot data fit bimanual projections using moving foreground regions and URDF geometry. UMI data extract device candidates in the tabletop region and optimize camera parameters using correspondences between tracked poses and images.}
\label{fig:camera-calibration}
\label{fig:calibration-evidence}
\end{figure}

\subsection{Quality Checks}
\label{sec:offline-quality-control}

Quality checks cover signal completeness, media quality, temporal consistency, and geometric validity.
Checks are enabled according to each source's available fields and embodiment characteristics, producing per-frame flags and trajectory-level statistics.
Table~\ref{tab:data-checks} lists checks in execution order; thresholds are source-specific.

Spike detection combines smoothing residuals with second- and third-order differences.
A spike is flagged when the residual and at least one higher-order difference exceed their thresholds.
Robust scales are estimated separately for each signal, with source-specific minimum thresholds for frame-to-frame jumps.
Extreme-value detection uses per-trajectory quantile tolerance bands; out-of-band samples are flagged as soft anomaly candidates.
Coordinate conventions are stored as source-level metadata for subsequent unification of reference frames and end-effector axes.

\begin{table}[tbp]
\centering
\caption{\textbf{Offline quality checks, listed in execution order.} Checks and thresholds are configured by data source.}
\label{tab:data-checks}
\tablefont
\begin{tabularx}{\linewidth}{lX}
\toprule
Check & Content and criteria \\
\midrule
Action completeness & Dimensions, value ranges, frame-to-frame increments, and constant signals \\
Media quality & Video readability, image dimensions, mean grayscale intensity, dark-pixel proportion, and sharpness \\
Geometric visibility & Camera-frame end-effector depth and image-boundary constraints \\
Extreme values & Per-trajectory tolerance band $[q_{.01}-\alpha\Delta q,\ q_{.99}+\alpha\Delta q]$, where $\Delta q=q_{.99}-q_{.01}$; out-of-band values are soft candidates \\
Spikes & Smoothing residuals and second-/third-order differences assessed against their respective robust scales; both the residual and at least one higher-order difference must exceed thresholds \\
End-effector continuity & Linear and angular velocities between adjacent poses, and rotation representation validity \\
Stationary and frozen signals & Frame-to-frame motion and duration identify stationary boundaries, internal pauses, and repeated state vectors \\
Scene motion & Frame-to-frame changes in low-resolution images, combined with action stationarity, identify stationary scene segments \\
State--action alignment & Cross-correlation estimates lag; after compensation for diagnostic purposes, difference signs are compared on moving frames. State-readback and low-motion dimensions are skipped \\
Rotation validity & Unit-norm checks for end-effector quaternions; full forward-kinematics consistency is outside the current check scope \\
Coordinate conventions & World orientation, base frame, quaternion ordering, and end-effector axis definitions \\
\bottomrule
\end{tabularx}
\end{table}

\FloatBarrier
\subsection{Filtering, Repair, and Validity Masks}
\label{sec:training-validity-masks}

Offline checks affect training through five mechanisms: quality flags, trajectory filtering, signal repair, anchor exclusion, and future-timestep masking.
Each is configured by source; not every source uses all five.

\begin{figure}[!htbp]
\centering
\includegraphics[width=\linewidth]{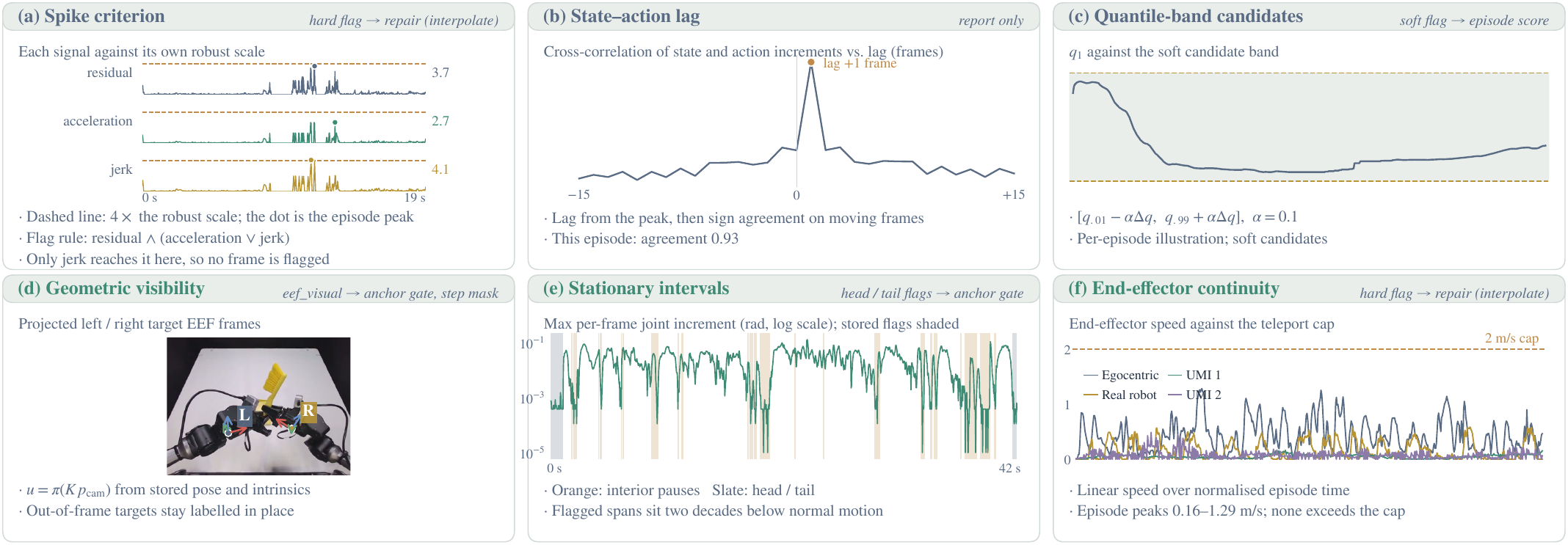}
\caption{\textbf{Quality-check examples.} (a) Smoothing residuals and second-/third-order differences relative to their respective robust scales. (b) State--action lag. (c) Quantile tolerance bands. (d) End-effector projections. (e) Frame-to-frame joint increments and flagged stationary intervals. (f) End-effector linear velocity. The $4\times$ and \SI{2}{m/s} thresholds are illustrative references; actual criteria depend on the source. The right side of each panel's title bar indicates where its output is used. Spikes and end-effector discontinuities are hard flags repaired by interpolation where enabled. Quantile bands produce soft flags used only in trajectory-level scoring. State--action lag is diagnostic only. End-effector visibility and boundary stationarity flags enter anchor gating and timestep masks.}
\label{fig:data-quality}
\end{figure}

\paragraph{Trajectory filtering and boundary trimming.}
Deletion occurs only at trajectory level or at trajectory boundaries.
Entire trajectories are discarded if shorter than 60 frames, if joint values exceed physical limits, or if retargeting fails.
Egocentric human data additionally trim flagged stationary segments at the beginning and end.
Criteria include boundary stationarity and redundant segments where both the scene and proprioceptive signals remain stationary.
Internal frames are retained; pauses receive flags, with their use determined by anchor gating below.

\paragraph{Signal repair.}
Repair is configured by source.
For egocentric human data and real-robot sources with repair enabled, frames flagged for spikes or end-effector discontinuities are linearly interpolated from neighboring valid frames.
Rotation components are renormalized.
Signals from sources without repair enabled are not modified.
Other quality flags remain associated with trajectories for use during training.

\paragraph{Anchor exclusion.}
During training, current-frame quality flags determine whether a frame can serve as an anchor.
Sources with anchor gating use combinations of image availability, end-effector visibility, and boundary stationarity flags.
Sources without gating retain quality flags with the data.

\paragraph{Future-timestep masking.}
Future action quality flags determine whether corresponding timesteps contribute to the action chunk loss.
Some sources use both anchor gating and future-timestep masks; others use only anchor gating or retain quality flags without applying them.
Anchor validity and individual timestep validity are independent, so an anchor may supervise only part of a chunk.
Figure~\ref{fig:data-quality} illustrates representative checks and uses of quality fields.

\section{World-Representation Targets}
\label{app:targets}

This section describes extraction of the three world supervision targets.
Their teachers, native resolutions, and layouts differ; retaining these formats would prevent readout from a common set of spatial queries.
We therefore rearrange them onto a common spatial grid.
Geometry and motion can then share 3D hidden states, while semantics aligns on a grid with the same structure.
The three losses in Equation~\eqref{eq:total-objective} can consequently use a common indexing scheme.
Frozen teachers produce
\begin{equation}
\begin{aligned}
\mathbf Y_t^{\mathrm{geo}}&=R_{\mathrm{geo}}\!\left(
f_{\mathrm{DA3}}^{\mathrm{T4W}}(\obs_t,\obs_{t+\Delta})_{\mathrm{src}}\right),\\
\mathbf Y_{t\rightarrow t+\Delta}^{\mathrm{mot}}&=R_{\mathrm{mot}}\!\left(
h_{\mathrm{3DFlow}}^{\mathrm{T4W}}(\obs_t,\obs_{t+\Delta})\right),\\
\mathbf Y_{t+\Delta}^{\mathrm{sem}}&=f_{\mathrm{DINO}}(\obs_{t+\Delta}),
\end{aligned}
\label{eq:teacher-targets}
\end{equation}
where $R_{\mathrm{geo}}$ and $R_{\mathrm{mot}}$ are deterministic spatial rearrangement operators.
Track4World processes each frame pair once, and geometry targets use source-frame features.
$R_{\mathrm{geo}}$ uses area resampling to convert an $18\times18$ grid to $16\times16$.
$R_{\mathrm{mot}}$ converts $32\times32$ motion features to $16\times16\times1024$ through $2\times2$ space-to-channel rearrangement.
DINOv3 natively produces $16\times16$ patch tokens from $256\times256$ inputs.
Teacher parameters remain frozen throughout training and are excluded from policy checkpoints.
Samples with non-finite motion targets are retried individually; those remaining invalid are excluded through teacher validity masks.

\section{Loss Definitions}
\label{app:losses}

This section supplements the main text with flow-time sampling and mask normalization for each loss.
Training linearly interpolates normalized action chunks with Gaussian noise:
\begin{equation}
\mathbf x_\tau=(1-\tau)\action+\tau\bm{\epsilon},\qquad
\bm{\epsilon}\sim\mathcal{N}(\mathbf 0,\mathbf I),\qquad
\mathbf u^\star=\bm{\epsilon}-\action.
\label{eq:flow-path}
\end{equation}
Flow time is sampled as $\tau=0.001+0.998b$, where $b\sim\operatorname{Beta}(1.5,1)$.
The masked mean squared error in Equation~\eqref{eq:flow-objective} is normalized by the number of valid elements.
This prevents action dimensionality or valid timestep counts from directly determining the loss scale.
Its full form is
\begin{equation}
\Loss_{\mathrm{flow}}=
\mathbb{E}_{\action,\bm{\epsilon},\tau}\left[
\frac{\|\mathbf M_{\mathrm{act}}\odot(\mathbf v_\theta(\mathbf x_\tau,\tau)-\mathbf u^\star)\|_2^2}
{\max(1,\|\mathbf M_{\mathrm{act}}\|_1)}\right].
\label{eq:flow-objective-masked}
\end{equation}
The semantic loss uses cosine distance:
\begin{equation}
\Loss_{\mathrm{sem}}=
\frac{\sum_i m_i^{\mathrm{sem}}\left[1-\cos(\hat{\mathbf z}_i^{\mathrm{sem}},\mathbf y_i^{\mathrm{sem}})\right]}
{\max(1,\sum_i m_i^{\mathrm{sem}})}.
\end{equation}
Geometry and motion losses use mean squared error after parameter-free channel normalization:
\begin{equation}
\Loss_r=
\frac{\sum_i m_i^{\mathrm{3d}}\|\operatorname{LN}(\hat{\mathbf z}_i^r)-\operatorname{LN}(\mathbf y_i^r)\|_2^2/d}
{\max(1,\sum_i m_i^{\mathrm{3d}})},
\qquad r\in\{\mathrm{geo},\mathrm{mot}\},\quad d=1024.
\end{equation}
$m_i^{\mathrm{sem}}$ indicates whether the corresponding view is valid at both current and future frames.
$m_i^{\mathrm{3d}}$ extends the sample-level teacher validity mask to individual tokens.
$\operatorname{LN}$ is applied to both student and teacher features.
$\Loss_{\mathrm{VQA}}$ computes autoregressive cross-entropy over answer tokens with valid question--answer annotations.
$\Loss_{\mathrm{FAST}}$ computes autoregressive cross-entropy over valid discrete action tokens.

\section{Model and Training Configuration}
\label{app:pretrain-config}

Table~\ref{tab:model-config} summarizes the principal model specifications, and Table~\ref{tab:training-config} lists pre-training optimization and execution settings.

\begin{table}[H]
\centering
\caption{\textbf{Principal configuration of \model.}}
\label{tab:model-config}
\tablefont
\renewcommand{\arraystretch}{1.02}
\begin{tabularx}{\linewidth}{lX}
\toprule
Configuration & Setting \\
\midrule
Total model parameters & 3.4B \\
VLM backbone & Qwen3.5-2B (2.2B parameters) \\
Expert hidden dimension & 1024; feed-forward intermediate dimension 3072 \\
Aligned depth & 24 layers: 18 within-stream layers and 6 joint attention layers (4, 8, \ldots, 24) \\
Within-stream layers & Gated DeltaNet: 16 key heads and 16 value heads, head dimension 128, convolution kernel size 4 \\
Joint attention layers & 8 query heads, 2 key--value heads, head dimension 256 \\
World queries & Semantic stream: up to $3\times256$; 3D stream: 256 \\
Spatial embeddings & $16\times16$ per view; content plus row/column positional embeddings, with additional camera embeddings for the semantic stream \\
Teachers & Geometry and motion: Track4World; semantics: DINOv3 ViT-L/16 \\
Teacher targets & 256 tokens per view, each with 1024 dimensions \\
Image inputs & $256\times256$, resized with preserved aspect ratio and centered zero-padding \\
States and actions & 34 unified slots, with validity masks by dimension and timestep \\
Action chunk length & $H=50$ \\
Prediction horizon & $\Delta_i=\lceil H/\rho_i\rceil$, aligned using the source-specific action sampling multiplier \\
Discrete action supervision & FAST action tokenizer \\
Inference integration & Euler, uniform step size, 10 steps by default \\
\bottomrule
\end{tabularx}
\end{table}

\begin{table}[htbp]
\centering
\caption{\textbf{Pre-training optimization and execution settings.}}
\label{tab:training-config}
\tablefont
\renewcommand{\arraystretch}{1.02}
\begin{tabularx}{\linewidth}{lX}
\toprule
Item & Setting \\
\midrule
Optimizer & Muon, with an AdamW branch for non-matrix parameters \\
Learning rate & $2\times10^{-4}$; VLM learning rate multiplied by 0.1 \\
Weight decay / gradient clipping & 0.01 / 1.0 \\
Learning rate schedule & 5\% warmup, then cosine decay to 0.1 of the peak \\
Execution & Distributed data parallel (DDP), BF16, gradient checkpointing, FlashAttention-2 \\
Action loss weight & 1.0 \\
Vision--language cross-entropy weight & 0.1, without scheduling \\
FAST weight & $0.05\rightarrow0.01$ \\
Semantic weight & $0.15\rightarrow0.05$ \\
Outer 3D weight & $0.25\rightarrow0.05$; inner geometry/motion weights $0.5/1.0$ \\
\bottomrule
\end{tabularx}
\end{table}

The outer weights $\lambda_{\mathrm{FAST}}$, $\lambda_{\mathrm{sem}}$, and $\lambda_{\mathrm{3d}}$ are scheduled by training progress $p=g/G$.
Here, $g$ is the current step and $G$ the total number of training steps.
Weights remain at their maximum for the first 20\% of training, undergo cosine decay over the middle 60\%, and remain at their minimum for the final 20\%:
\begin{equation}
\lambda(p)=
\begin{cases}
\lambda_{\max}, & p\leq 0.2,\\[2pt]
\lambda_{\min}+(\lambda_{\max}-\lambda_{\min})\cdot
\tfrac{1}{2}\left[1+\cos\!\left(\pi\tfrac{p-0.2}{0.6}\right)\right], & 0.2<p<0.8,\\[4pt]
\lambda_{\min}, & p\geq 0.8.
\end{cases}
\label{eq:aux-weight-schedule}
\end{equation}
The action loss weight and inner geometry/motion weights remain fixed; Table~\ref{tab:training-config} lists all values.
Vision--language batches execute only the VLM forward pass.
Their autoregressive cross-entropy contributes to the total objective with a fixed weight of $0.1$, independent of the auxiliary weight schedule.

\endgroup
\end{document}